\documentclass[journal,twoside,web]{ieeecolor}
\usepackage{jsen}
\usepackage{cite}
\usepackage{amsmath,amssymb,amsfonts}
\usepackage{algorithmic}
\usepackage{graphicx}
\usepackage{textcomp}
\usepackage{wrapfig}
\usepackage{algorithm}
\usepackage{algorithmic}
\usepackage{bm}
\if CLASSOPTIONcompsoc
\usepackage[caption=false,font=normalsize,labelfont=sf,textfont=sf]{subfig}
\else
\usepackage[caption=false,font=footnotesize]{subfig}
\fi

\def\BibTeX{{\rm B\kern-.05em{\sc i\kern-.025em b}\kern-.08em
    T\kern-.1667em\lower.7ex\hbox{E}\kern-.125emX}}
\definecolor{abstractbg}{rgb}{0.89804,0.94510,0.83137}
\begin{document}
\title{Robust and accurate depth estimation by fusing LiDAR and Stereo}
\author{Guangyao~Xu,
	Junfeng~Fan,
%	\IEEEmembership{Member, IEEE}
	En~Li, 
%	\IEEEmembership{Member, IEEE}
	Xiaoyu~Long,
	and~Rui~Guo
\thanks{This work was supported in part by the National Key Research and
	Development Program of China under Grant 2018YFB1307400 and in part by
	the National Natural Science Foundation of China under Grant 61873267.}% <-this % stops a space
\thanks{Guangyao~Xu, Junfeng~Fan, En~Li, and Xiaoyu~Long, are with the School of Artificial Intelligence, University of the Chinese Academy of Sciences, No.19(A) Yuquan Road, Beijing, 100049, China and also with the Key Laboratory of Complex Systems and Intelligence Science, Institute of Automation, Chinese Academy of Sciences, Beijing, 100190, China.  e-mail: (xuguangyao2019@ia.ac.cn; fanjunfeng2014@ia.ac.cn; en.li@ia.ac.cn; longxiaoyu2019@ia.ac.cn).}
\thanks{Rui~Guo is with Electric Power Research Institute, State Grid Shandong Electric Power Company, Jinan 250003, China.
	e-mail: (guoruihit@qq.com).}
}

\IEEEtitleabstractindextext{%
\fcolorbox{abstractbg}{abstractbg}{%
\begin{minipage}{\textwidth}%
\begin{wrapfigure}[12]{r}{2.8in}%
\includegraphics[width=2.7in]{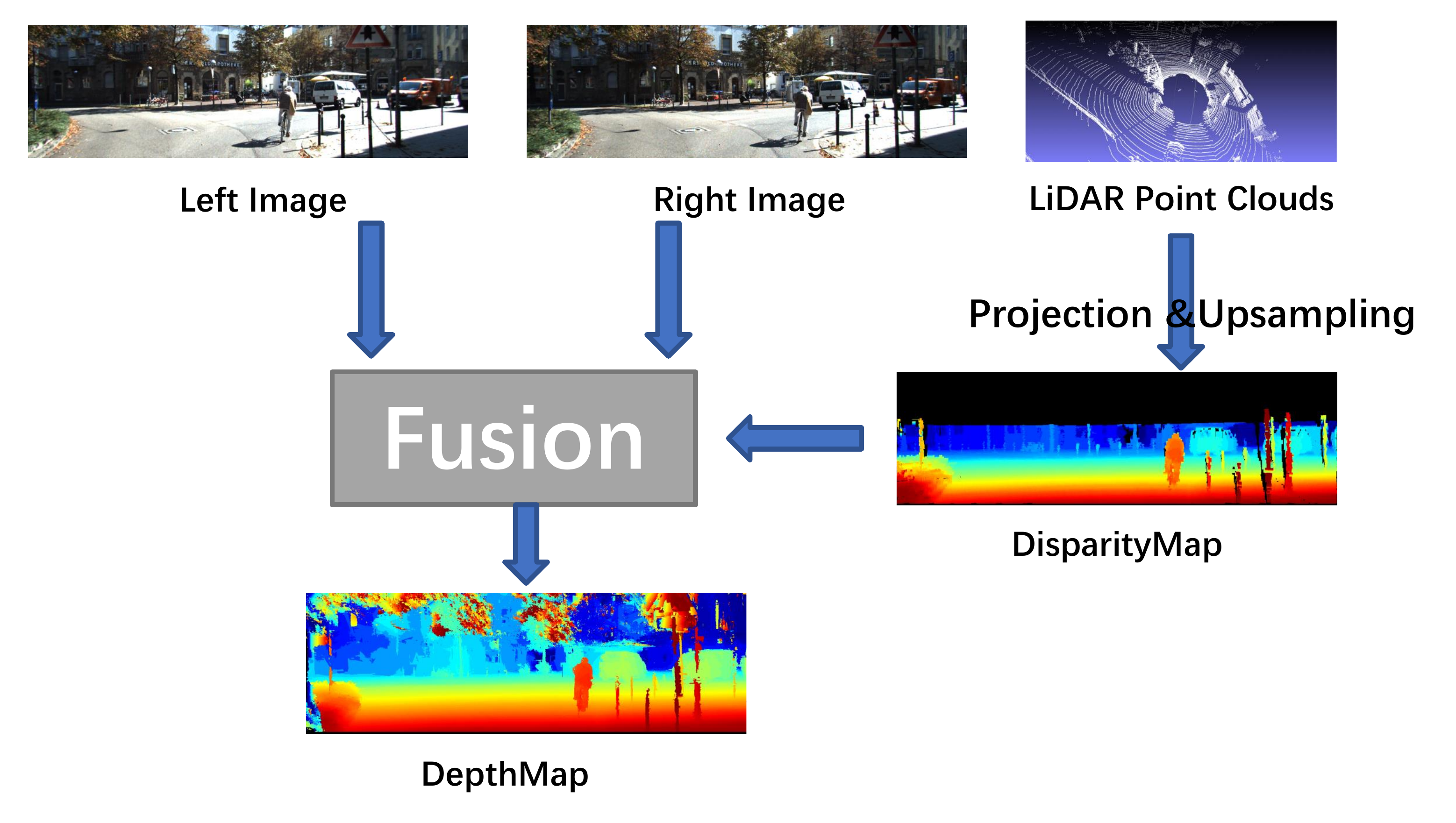}%
\end{wrapfigure}%
\begin{abstract}
	Depth estimation is one of the key technologies in some fields such as autonomous driving and robot navigation.
However, the traditional method of using a single sensor is inevitably limited by the performance of the sensor. Therefore, a precision and robust method for fusing the LiDAR and stereo cameras is proposed. This method fully combines the advantages of the LiDAR and stereo camera, which can retain the advantages of the high precision of the LiDAR and the high resolution of images respectively. Compared with the traditional stereo matching method, the texture of the object and lighting conditions have less influence on the algorithm. 
%that is suitable for use in large outdoor scenes.
%	First, LiDAR data can be projected onto stereo images as a sparse disparity map that is converted from a depth map.
Firstly, the depth of the LiDAR data is converted to the disparity of the stereo camera. 
Because the density of the LiDAR data is relatively sparse on the y-axis, the converted disparity map is up-sampled using the interpolation method.
%Compared with the bilateral filtering method, the linear interpolation has a higher accuracy when the density of the LiDAR scan line is relatively high.
Secondly, in order to make full use of the precise disparity map, the disparity map and stereo matching are fused to propagate the accurate disparity. 
Finally, the disparity map is converted to the depth map.
%	Patch Match Stereo (PMS) method is used to spread disparity with the lower matching cost during the stereo matching process, reducing the overall image matching cost. And the slanted support windows of PMS normally have higher accuracy compared to the traditional front-parallel window. 
Moreover, the converted disparity map can also 
%	restrict the search range of disparity
%between the scan lines of LiDAR to reduce the search range 
increase the speed of the algorithm.
%	The traditional stereo matching algorithm has higher accuracy, robustness than before, because of the addition of LiDAR data.
We evaluate the proposed pipeline on the KITTI benchmark. 
The experiment demonstrates that our algorithm has higher accuracy than several classic methods.
\end{abstract}

\end{minipage}}}

\maketitle

\section{INTRODUCTION}
\label{sec:introduction}
\IEEEPARstart{T}{he} stereo camera is susceptible to interference from external environmental factors, such as light conditions. Multiple camera systems can get more information from different points of view, but these systems don't improve the anti-interference ability. Contrary to cameras, LiDARs have high anti-interference performance, which can obtain accurate depth information in complex environments, but the data of LiDARs is dense on the x-axis, but the low resolution on the y-axis. In addition to the stereo camera and LiDAR, RGB-D cameras are also often used to measure depth information, such as Kinect\cite{kinect} and Realsense \cite{keselman2017intel}. But the RGB-D camera is generally suitable for measuring short-distance targets. Moreover, sunlight can interfere with its structured light sensor, so it is usually difficult to be used in a outdoor environment. In general, each sensor has its own strengths and limitations. The strategy of multi-sensor fusion is to make full use of the advantages of different sensors and overcome their limitations. At present, most of the fusion works of stereo and LiDAR are to integrate the data of the stereo camera and the 3D LiDAR in different ways to achieve an accurate estimation of the depth\cite{premebida2016high} \cite{maddern2016real} \cite{shivakumar2019real} \cite{wang20193d} \cite{choe2021volumetric} \cite{zhang2020listereo}. 
% 	Because depth estimation plays an important role in many fields such as autonomous driving and robotics.
Although recently scholars have proposed many new stereo and LiDAR fusion methods and achieved good results, the new methods almost used deep learning methods which use massive data. But the lack of experimental data and the limitations of computing equipment prompted us to study new stereo camera and LiDAR fusion methods.

In the process of fusion of the stereo camera and LiDAR, the positional relationship between LiDAR and stereo camera can be needed. It builds a bridge between different sensors to facilitate the acquisition of the correspondence between point clouds and images. According to statistics, after the LiDAR point clouds are projected onto the pixel plane, they only occupy about 6.8\%\cite{scharstein2003high} of the image pixel position. In addition, the depth information of the LiDAR point cloud also can be converted into the accurate disparity information of the stereo camera. If the disparity converted from the point clouds is not in the pixel plane of the stereo camera, it would be discarded. And the LiDAR point clouds can help us to reduce the possible search range of the disparity, thereby reducing the time required for the calculation. In order to make full use of this precise disparity information, it is up-sampled. Then the precise disparity can be combined with stereo matching to find a suitable position by means of propagation. 

The main contributions of our paper are: 
Firstly, a new method of removing outliers is proposed.
%	Firstly, a novel upsampling method of lidar disparity map is proposed. Compared with $BF^{*}$ \cite{premebida2016high}, our method $BI^{*}$ has better accuracy and robustness. 
Secondly, a new pipeline of fusion of the stereo images and raw LiDAR data is proposed.

An experimental platform was built to verify the feasibility of our method. And we also conducted lots of experiments on the KITTI benchmark to evaluate the efficiency of the proposed method.

\section{RELATED WORK}

In order to obtain reliable depth information, many methods have been proposed. In recent years, these methods are mainly concentrated on three aspects, for example, the stereo matching of stereo images, stereo camera and LiDAR data fusion and the completion of depth maps. 
%	deep learning
\subsection{Stereo matching}
MC-CNN \cite{zbontar2016stereo} is one of the earliest methods to use deep learning for stereo matching. And it shows outstanding performance on the KITTI benchmark and proves the good performance of deep learning neural networks in the field of stereo matching. 
Garg et al. \cite{garg2016unsupervised} creatively used an unsupervised deep learning method to estimate depth, because the ground truth is difficult to obtain. This method uses the photometric error as the loss function. But the results of this algorithm are a little fuzzy.
Next, Godard et al. \cite{godard2017unsupervised} added left-right consistency to the unsupervised approaches. Compared with existing unsupervised ways, their method effectively improves robustness and performance.
Deep learning has made great progress in the field of vision. It always occupies the top few in data sets such as the KITTI benchmark \cite{kitti}. However, supervised learning methods' \cite{hastie2009overview} dependence on data and computing power makes it difficult to apply. 
Compared with supervised learning methods, unsupervised learning methods \cite{barlow1989unsupervised} do not rely on ground truth and thus can be applied better. 
\subsection{LiDAR and stereo fusion}
In the field of multi-sensor fusion, several methods have been also proposed. 
In \cite{shivakumar2019real}, Shivakumar et al. performed anisotropic diffusion of depth data in the cost space of the SGM method. It can be run in real-time on the TX2 platform.
In \cite{maddern2016real}, Will Maddern and Paul Newman proposed to use probabilistic methods to fuse LiDAR and stereo cameras. The performance of this method may degrade for the regions that LiDAR data is missing. 
Compared with traditional methods, Park et al. \cite{park2018high} used deep learning methods for depth estimation with LiDAR and stereo camera. They designed the two-step cascade deep architecture, consisting of a disparity fusion module and disparity refinement module. The features were extracted from LiDAR data and stereo disparity, which is fused in the LiDAR-Stereo fusion module. Then the fused disparity and RGB image were fused to estimate the real depth data in the refinement network. This method straightly uses stereo disparity maps. But they can be refined by LiDAR data in the process of stereo matching. The input data should be as accurate as possible.
Wang et al. \cite{wang20193d} introduced the Input fusion and CCVNorm into the stereo and lidar fusion network and achieve good results.
The Input fusion combines the lidar pair and the image pair as a fusion layer to learn the joint feature representation.
The CCVNorm adjusts the cost volume features based on lidar data.
Choe et al. \cite{choe2021volumetric} proposed a network that includes feature extraction, volume aggregation, and depth regression stage. The feature extraction stage extracts feature points and feature vectors of images and LiDAR data. The volume aggregation stage combines these features and calculates the geometric cost. Finally, the depth map is calculated in the depth regression stage.
The methods \cite{park2018high,wang20193d,choe2021volumetric} need extract features from lidar data. But they may not contain enough features.
The approach we proposed makes LiDAR data improve the results of stereo matching and doesn't have any special requirements for lidar information. 
\subsection{Depth completion}
Some scholars propose to directly use raw LiDAR data for depth completion. We mainly focused on depth completion using traditional methods.
Badino et al. \cite{huber2011integrating} proposed the depth map up-sampling method. The general process of this approach is to first sample in the horizontal direction and then sample in the vertical direction. 
%	They reduce the calculating work by calculating the maximum and minimum disparity to shorten the time. 
Premebida et al. \cite{premebida2016high} used bilateral filtering for up-sampling. In Fig. \ref{fig3:a}, moreover, the LiDAR data of different depths are projected onto the same region, especially in the edge area. In order to eliminate the error points, they used DBSCAN \cite{sander1998density} method to cluster the LiDAR data in terms of the different depths and then retain the data with the correct depth value. But this method is very time-consuming. And lidar data is smoothed that would change data characteristics.
%	This approach has a significant improvement compared to bilateral filtering. 
Ku et al. \cite{ku2018defense} performed morphological processing on the depth image and achieved good up-sampling results. This method runs in real-time. But this method has the same disadvantage with \cite{premebida2016high}.
In addition to the above mentioned, Hengjie et al. \cite{9450836} proposed a deep learning method for depth estimation using single-line lidar and an image.
Compared with multi-line lidar, the single-line lidar has the advantages of simple structure and low price. But this is an ill problem in math. 
%	The main contribution of our paper are: Firstly, a novel up-sampling method of lidar disparity map is proposed. Compared with $BF^{*}$ \cite{premebida2016high}, our method $BI^{*}$ has better accuracy and robustness.
%	Secondly, a pipeline of fusion of the stereo camera and raw LiDAR data is proposed. 
%	
%	Both sensors have some advantages to complement each other. For example, the camera image has a higher resolution and color information, while the LiDAR data has accurate depth information. Compared with other methods, the reliable depth information provided by LiDAR is used to correct the stereo matching process by slanted support windows and improve the accuracy and robustness of the stereo matching algorithm.
% 	The proposed approach fused LiDAR data and stereo images for depth estimation based on PMS. It can achieve good results on the Middlebury \cite{scharstein2003high} data set, but the effect on the KITTI data set is not good, mainly because the lighting and texture of the scenes collected outdoors are bad. Our proposed method can greatly improve the accuracy and speed of PMS at the same time. Fig.\ref{fig1} shows the input data and the final result of our pipeline.

The rest of the article is organized as follows. In section 3, the disparity-map up-sampling method is proposed. Section 4 introduces the detailed process of depth estimation. Section 5 is some experiments and data analysis. Section 6 is some summary and improvement measures.

\section{DISPARITY-MAP UPSAMPLING FORMULATION}

\begin{figure} %%跨栏图为：figure*而单栏图为 figure
	\centering
	\subfloat{ \includegraphics[width=3.5in]{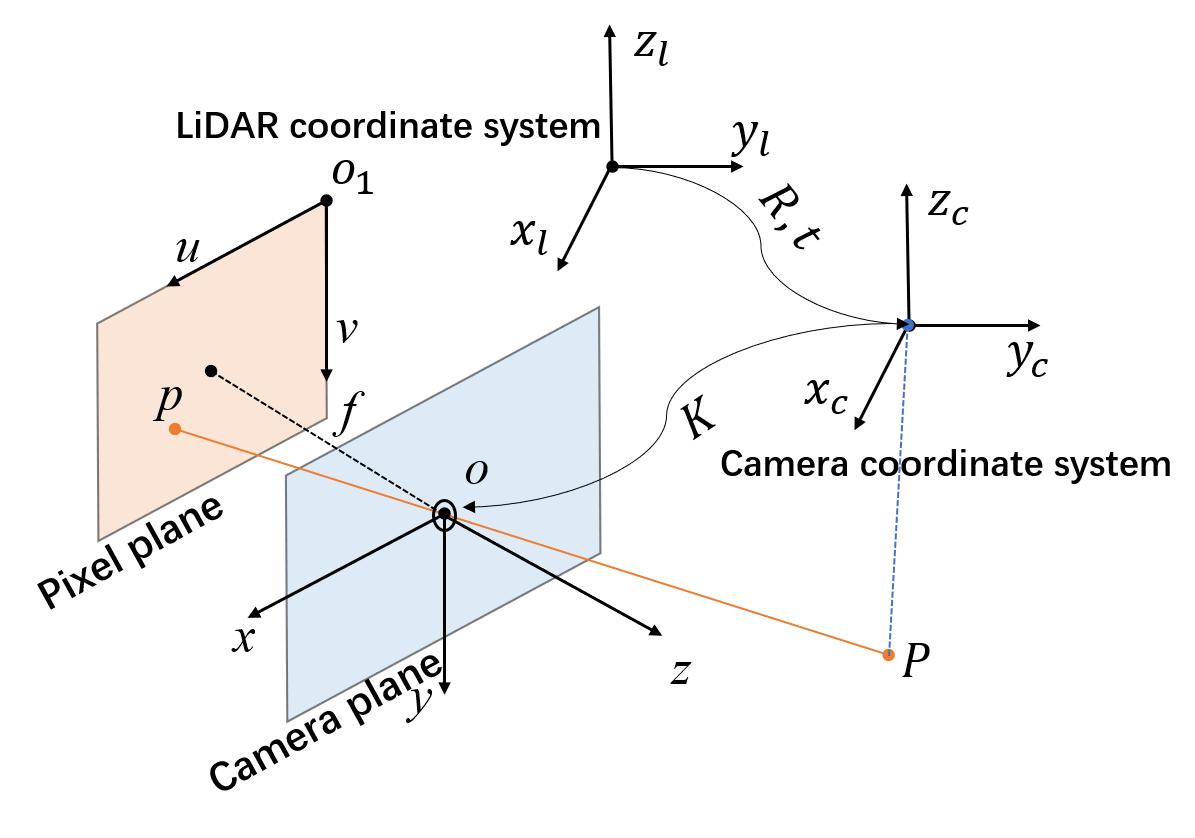} }	
	\caption{The process of projecting LiDAR points to the pixel plane.}
	\label{coordinate}
\end{figure}  %%跨栏图为：figure*而单栏图为 figure
In order to fuse LiDAR data and stereo images, the depth of the lidar is converted into the disparity and then up-sampled. 
The disparity map up-sampling and depth completion are the same tasks in this article.
%	And then our method requires a dense disparity map as an important data item for calculating the matching cost in the next section. 
%	The pipeline needs to input a pair of the rectified stereo image and LiDAR data. 
The pair of stereo images is divided into left image $L$ and right image $R$. 
In addition, the LiDAR data corresponding to the stereo images are projected to the image plane of $L$ denoted by $\Pi_{l}$ or the image plane of $R$ denoted by $\Pi_{r}$ as the disparity map according to Eq.\ref{lidar} and Eq.\ref{disparity}. We assume that the stereo camera is a parallel optical axis system.
\begin{align}
	Y=K[R|t]X
	\label{lidar}
\end{align}
where $X=(x^{'},y^{'},z^{'},1)^{T}$ denotes a 3D LiDAR point. And $Y=(x,y,1)^{T}$ denotes a pixel in the camera image, the position of which may not be an integer pixel coordinate. So it can be rounded.  
\begin{equation}
	K = \begin{bmatrix}
		f_{x}&0&c_{x}&0\\
		0&f_{y}&c_{y}&0\\
		0&0&1&0\\
	\end{bmatrix}
\end{equation}
$K$ is the camera projection matrix.
$(f_{x},f_{y})$ and $(c_{x},c_{y})$ are the focal length and principal point of stereo cameras.  
$R\in\mathbb{R}^{3*3}$ and $t\in\mathbb{R}^{3*1}$ are rotation and translation matrix from LiDAR to camera.
\begin{equation}
	[R|t] = \begin{bmatrix}
		R&t\\
		0&1\\
	\end{bmatrix}
\end{equation}
And then the depth $z^{'}$ of $X$ is converted to disparity $d$ at the pixel coordination $(x,y)$. 
\begin{align}
	d=\frac{BF}{z^{'}}
	\label{disparity}
\end{align}
where $B$ and $F$ are the baselines and the focal length of stereo cameras.
Fig. \ref{coordinate} shows the process of projection.
When the left and right images are not distinguished, we uniformly express the left image as $\Pi$ which is the baseline disparity map. The disparity map $\Pi$ can be considered as an 8bit image. The disparity in $\Pi$ is denoted as $d$. Although the points in $\Pi$ are very sparse, there are a few points of different depths projected to the same local area, such as the magnified area of Fig .\ref{fig3:a}. The blue points' depth is more than the red points.
%	In addition, the LiDAR cannot be seen as a point that emits light outward, which is located at the optical center of the lens. According to the camera's pinhole imaging model, this is the root cause of this phenomenon. 
This phenomenon is explained in the \cite{premebida2016high}. 
%Fig.\ref{fig2} intuitively explains this phenomenon. 
%\begin{figure}[!htbp] %%跨栏图为：figure*而单栏图为 figure
%	\centering
%	\subfloat{ \includegraphics[width=3.5in]{pic/fig2.pdf} }	
%	\caption{Points of different depths are projected to the same local area.}
%	\label{fig2}
%\end{figure}  %%跨栏图为：figure*而单栏图为 figure	

\begin{figure}[!htbp]
	\centering
	\subfloat{ \includegraphics[width=3.5in]{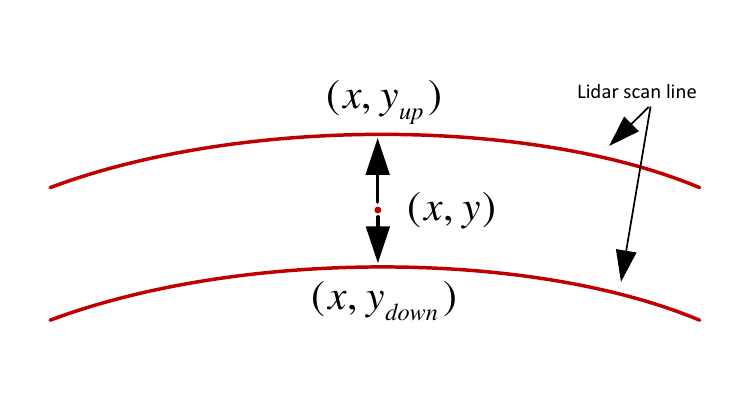} }	
	\caption{the pixels between scan lines.}
	\label{tujie}
\end{figure}

\begin{figure}[!htbp] %%跨栏图为：figure*而单栏图为 figure
	\centering
	\subfloat[LiDAR projection disparity map]{ \includegraphics[width=3.5in]{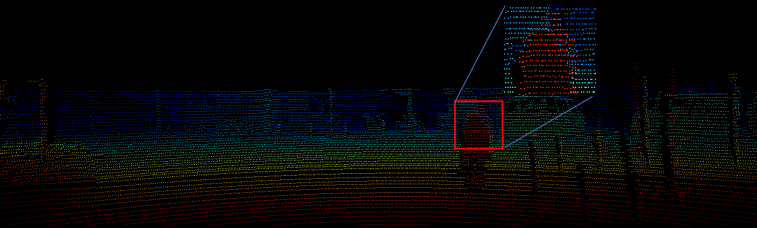} \label{fig3:a}}
	\quad
	\subfloat[Complement scan line]{ \includegraphics[width=3.5in]{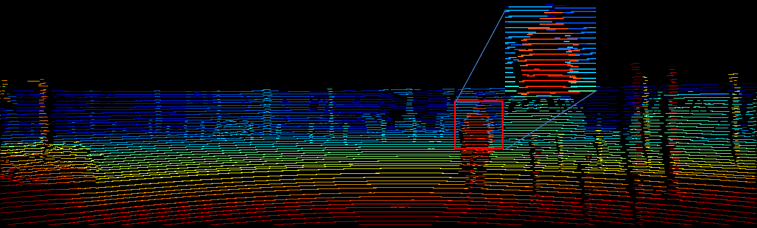} \label{fig3:b}}
	\quad
	\subfloat[Eliminate overlapping points]{ \includegraphics[width=3.5in]{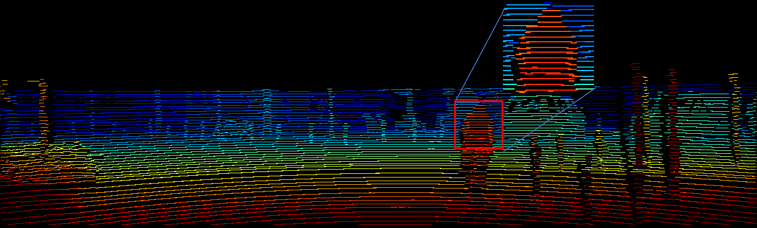} \label{fig3:c}}
	\caption{The process of eliminating outlier in the disparity map.}
	\label{fig3}
\end{figure}

\begin{figure}
	\centering
	\subfloat[linear interpolation on up-sampling]{ \includegraphics[width=3.5in]{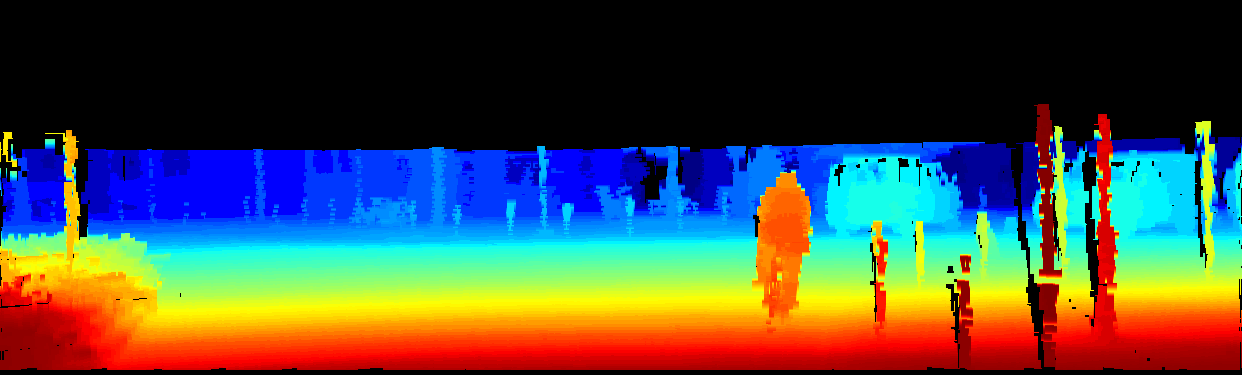} \label{BF:a}}
	\quad
	\subfloat[bilateral interpolation on up-sampling\cite{premebida2016high}]{ \includegraphics[width=3.5in]{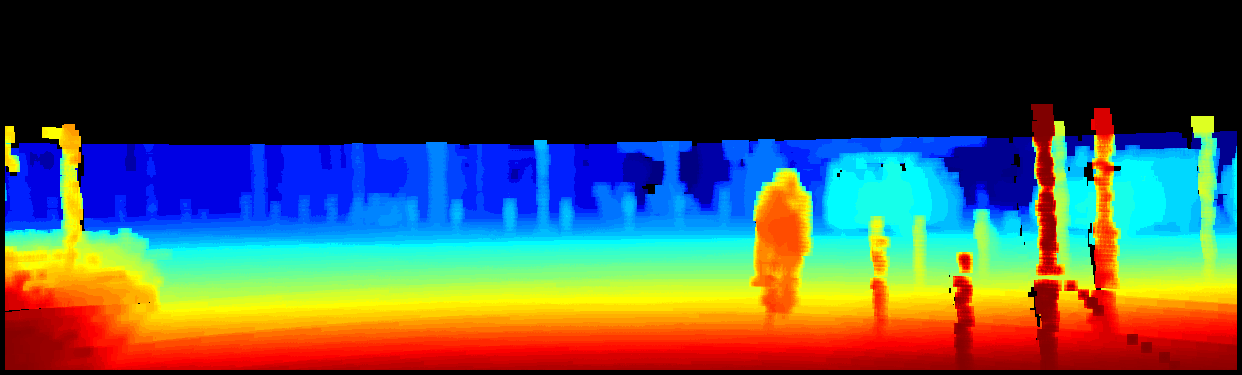} \label{BF:B}}
	\quad
	\caption{The result of up-sampling on the disparity map. (a) uses linear interpolation and (b) uses bilateral filtering of \cite{premebida2016high}.}
	\label{BF}
\end{figure}

This problem causes great trouble to our disparity map up-sampling. 
Algorithm 1 can be used to solve this problem to some extent which clears the outliers between scan lines. 
A brief summary of the algorithm 1 is as follows:
The disparity from LiDAR data has an obvious feature that is high resolution in the horizontal direction and low resolution in the vertical direction. The first step is to complete these scan lines in the horizontal direction and the result is shown in Fig.\ref{fig3:b}. 
And then the disparity coordination $(x,y)$ of $\Pi$ turn up and down to find the disparity coordination $(x,y_{up})$ and $(x,y_{down})$ like the Fig. \ref{tujie}. If the difference between $\Pi(x,y)$ and $\Pi(x,y_{up})$ ($\Pi(x,y_{down})$) is more than a special value, this disparity would be removed. 
%Since the LiDAR points are relatively sparse, the far points that should be blocked are projected near the near points. 
%After completing the scan lines, the far and close points all will become the scan lines.
%So a scan line that is farther away appears between two closer scan lines should be removed.  
\begin{algorithm}[!h]
	\caption{eliminate the overlapping points}
	\label {algorithm1}
	\begin{algorithmic}[1]
		\STATE Get projection disparity map
		\STATE Complement the scan line by interpolation method
		\STATE $UP\_FLAG \gets \FALSE$
		\STATE $DOWN\_FLAG \gets \FALSE$
		\FOR {the coordinate $(x_{d},y_{d})$ of each $d$ in $\Pi$}
		\FOR {$Yup$ $\gets$ $y_{d}$ \TO $y_{up}$}    
		\IF {$Yup > 0$ \AND $\Pi(x,Yup)-\Pi(x_{d},y_{d}) > Maxdisp$ \AND $Color(x,Yup)-Color(x_{d},y_{d}) < Maxcolor$}
		\STATE		$UP\_FLAG \gets \TRUE$
		\ENDIF
		\ENDFOR
		\FOR {$Ydown$ $\gets$ $y_{d}$ \TO $y_{down}$}    
		\IF {$Ydown<MaxY$ \AND $\Pi(x,Ydown)-\Pi(x_{d},y_{d}) > Maxdisp$ \AND $Color(x,Ydown)-Color(x_{d},y_{d}) < Maxcolor$}
		\STATE	$DOWN\_FLAG \gets \TRUE$
		\ENDIF
		\ENDFOR
		\IF {$UP\_FLAG=True$ \AND $DOWN\_FLAG=True$}
		\STATE $\Pi(x_{d},y_{d})=0$
		\ENDIF
		\ENDFOR
	\end{algorithmic}
\end{algorithm}
Fig.\ref{fig3:c} is the result of removing error points.
And then the y-axis of Fig.\ref{fig3:c} is linear interpolated between scan lines according to Eq. \ref{equ1}. 
%In other words, we interpolate between two LiDAR scan lines . 
\begin{align}
	\Pi(x,y)=&\dfrac{y-y_{up}}{y_{down}-y_{up}}\Pi(x,y_{down}) \notag \\
	&+\dfrac{y_{down}-y}{y_{down}-y_{up}}\Pi(x,y_{up})
	\label{equ1}
\end{align}
where $\Pi(x,y)$ is the disparity at $(x,y)$. $y_{up}$ and $y_{down}$ are y coordinate that are the closest to $(x,y)$ distributed on the scans lines.
%Fig.\ref{fig3} show the main steps of up-sampling. 
Fig.\ref{BF:a} shows the result after linear interpolation.
  %%跨栏图为：figure*而单栏图为 figure
Due to the fact that linear interpolation is not good at preserving the edge, some other scholars like \cite{premebida2016high} use bilateral filtering to up-sample disparity map. 
%	Since bilateral filtering is the sliding method of the window, it is progressively excessive at the boundary. According to the characteristics of LiDAR data, linear up-sampling can maintain the boundary in the vertical direction very well. 
In \cite{premebida2016high}, disparity map is up-sampled by bilateral filtering as shown in the Fig. \ref{BF:B}.
$BI^{*}$ represents the disparity map by the up-sampling method proposed in this paper, and $BF^{*}$ represents the up-sampling method proposed in \cite{premebida2016high}.
Compared with $BF^{*}$ \cite{premebida2016high}, our method $BI^{*}$ has better accuracy and robustness. 
In section 5, we compared the two up-sampling methods in detail.

\section{DEPTH ESTIMATION}
\label{sec:DEPTH ESTIMATION}
The most important step of depth estimation is cost calculation. However, due to environmental factors, the images captured by the camera lose part of the information. Moreover, LiDAR shows strong robustness to environmental noise. We add LiDAR data and look forward to obtaining more accurate matching costs to propagate the disparity.
%	 An up-sampled disparity map assigns an estimated value to most pixels. 
%	$D$ and $d$ can be denoted as disparity from LiDAR points and estimated disparity. 
\subsection{Window model}
\begin{figure}[!htbp] %%跨栏图为：figure*而单栏图为 figure
	\centering
	\subfloat{ \includegraphics[width=3.5in]{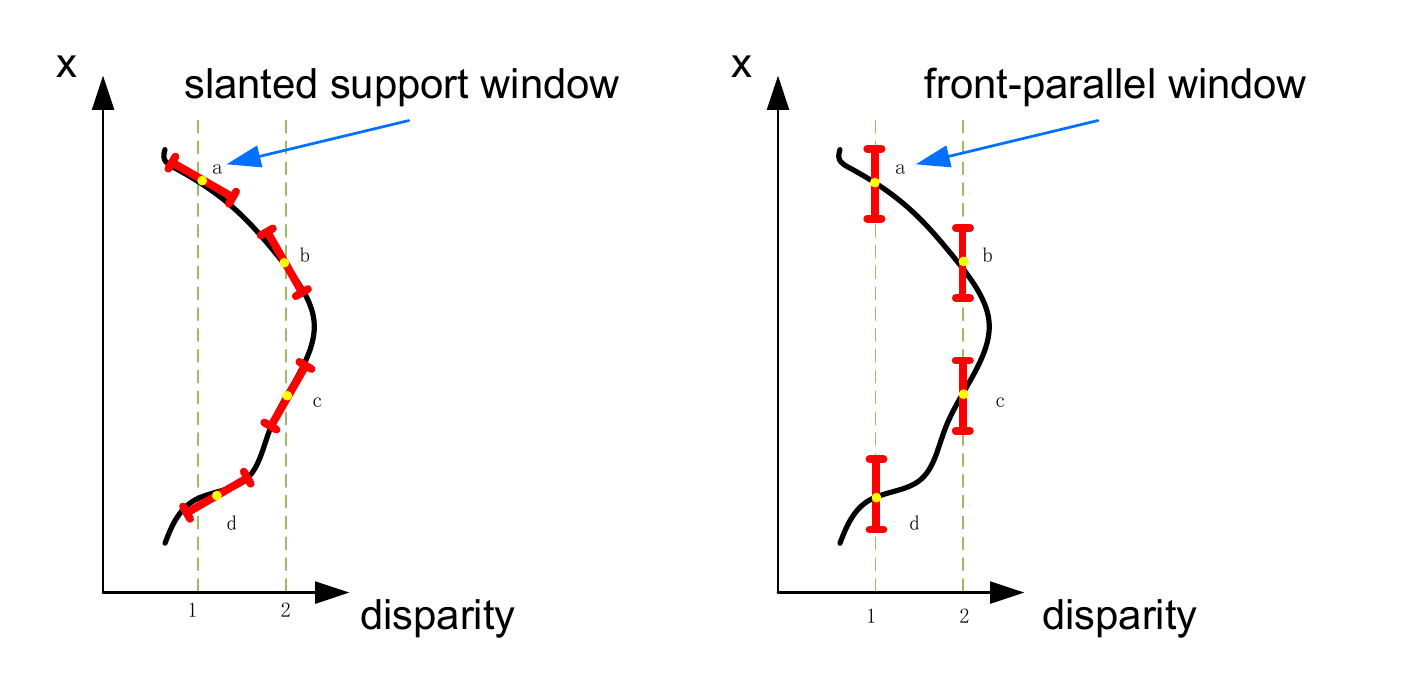} }
	\caption{Slanted support window and front-parallel window. The y-axis of the parallax plane is set to a specific value, so the three-dimensional scene is projected onto the two-dimensional image.}
	\label{fig4}
\end{figure}

%	We use slanted support windows instead of traditional parallel windows as shown in Fig. \ref{fig4}.
Front-parallel windows can be expressed as windows parallel to the image plane. 
The slanted support window \cite{bleyer2011patchmatch} can be denoted as the window that fits the disparity surface. Moreover, the disparity of front-parallel windows is integer disparity, and the slanted support window is based on sub-pixel disparity. For instance, slanted support window $d$ of Fig. \ref{fig4} is sub-pixel disparity between disparity $1$ and $2$. 
There is a significant difference between the slanted support window and the front-parallel window proposed as shown in Fig. \ref{fig4}. 
In section 5, We compare the effects of the two window models in detail.
  %%跨栏图为：figure*而单栏图为 figure
Slanted support window is used to allocate a plane $\xi$ with the least matching cost for each pixel $p$. The plane $\xi$ is determined by the point $ (x,y,d) $ and a random unit normal vector $\vec{n}=(n_{x},n_{y},n_{z})$ of the point.  
The plane $\xi_{p}$ can be represent as
\begin{align}
	d=a_{\xi}x+b_{\xi}y+c_{\xi}
\end{align}
where $(x,y,d)$ is a point of the disparity plane and the $(x,y)$ is pixel coordination in the disparity map. $a_{\xi}$, $b_{\xi}$ and $c_{\xi}$ are the three parameters of the plane $\xi$. These parameters are obtained by the following three formulas.
\begin{align}
	a_{\xi}= -\dfrac{n_{x}}{n_{z}} 
\end{align}

\begin{align}
	b_{\xi}=- \dfrac{n_{y}}{n_{z}} 
\end{align}

\begin{align}
	c_{\xi}=\dfrac{n_{x}x+n_{y}y+n_{z}d}{n_{z}} 
\end{align}
There are infinitely many planes such as $\xi$, which prevent us to find all possible planes. 
In order to save time, the plane can be found in a search range where the optimal solution is most likely to appear.
The search range is set between the minimum and the maximum disparity of the whole image in PMS\cite{bleyer2011patchmatch}.
But it also can be set between scan lines after adding LiDAR data. 
Because the disparity obtained from the LiDAR data is sufficiently accurate, the disparity $d$ on the scan line can be regarded as the maximum or minimum disparity of the pixels between the scan line.
If the two scan lines both don't exist or only one scan line, the search space is set in the minimum and the maximum disparity of the whole image. 
Allocating a smaller search space for disparity can greatly shorten the calculation time.
In addition, the search space of the normal vector is $-180$° and $+180$°. If the normal vector is $0$, the slanted window can be transformed into the front-parallel window. 
%	In order to find the best matching point, we can only find the minimum aggregation cost in the search range.
\subsection{Matching costs}
After determining the window model, the matching cost also needs to be determined.
$p(x,y)$ can be denoted as a pixel coordinate of disparity in the disparity map for convenience. The aggregation cost of points $(p,d)$ on plane $\xi$ is
\begin{align}
	C(p,\xi_{p})= \sum^{}_{(p^{'},d^{'})\in W_{p}} \varphi(p,p^{'})\cdot \phi(p^{'},p^{'}-(a_{\xi}x+b_{\xi}y+c_{\xi}))
	\label{cost}
\end{align}
where $W_{p}$ is a square window centered on $p$. $(p^{'},d^{'})$ is pixel coordinate and disparity in the window $W_{p}$ of $p$.

The first item of Eq. \ref{cost} is the function $\varphi(p,p^{'})$ that is used to solve the edge-fattening problem \cite{yoon2005locally}. 
\begin{align}
	\varphi(p,p^{'})=e^{-\dfrac{\lVert I_{p}-I_{p^{'}} \rVert_{1}}{\delta}}	
\end{align} 
where $\delta$ is a custom parameter that controls the proportion of pixels of different colors in the original image. In this paper, the value of $\delta$ is 10. $\lVert I_{p}-I_{p^{'}} \rVert_{1}$ calculates the Manhattan Distance between color value of $p$ and $p^{'}$ in RGB space. The value of $ \varphi(p,p^{'})$ would be large if color is similar. 
%	And it is calculated on the plane of $\xi_{p}$.  

The Eq. \ref{dissimiliar} that calculates the matching cost between $p^{'}$ and $q$ is the last section of the Eq. \ref{cost}. $q$ is expressed as $p^{'}-(a_{\xi}x+b_{\xi}y+c_{\xi})$ and also the matching points of $p^{'}$ obtained by subtracting the disparity of the x-axis from $p^{'}$ in the other view.  
\begin{align} 
	\phi(p^{'},q)&=\alpha\cdot min(\lVert I_{p^{'}}-I_{q} \rVert_{1},\lambda_{col}) \notag \\
	&+\beta\cdot min(\lVert  \nabla I_{p^{'}}- \nabla I_{q} \rVert_{1} ,\lambda_{grad}) \notag \\
	&+\gamma \cdot min(\lvert \Pi_{p^{'}}- d^{'} \rvert ,\lambda_{disp})  %标准视差和估计视差之间插值
	\label{dissimiliar}
\end{align} 
Here, $\lVert I_{p^{'}}- I_{q} \rVert_{1}$ and $\lVert  \nabla I_{p^{'}}- \nabla I_{q} \rVert_{1}$ are the Manhattan Distance in RGB and gradient. $\lvert \Pi_{p^{'}}- d^{'} \rvert$ is the absolute difference of the disparity. $\Pi_{p^{'}}$ is disparity of $p^{'}$ in up-sampled disparity map. $d^{'}$ is the disparity of $p^{'}$ in-plane $\xi_{p}$. The color and gradient come from the original image and the disparity comes from the disparity map.

%	The disparity obtained from LiDAR is fused into PMS, which is one of main innovations.
The custom parameters $\alpha$, $\beta$ and $\gamma$ are used to balance the ratio between color, gradient and disparity. 
%If the LiDAR point cloud is dense and accurate, the proportion of disparity can be appropriately increased in the formula. 
%In other words, if the image quality is better, the proportion of colors can be increased. $\alpha$ and $\gamma$ can be set to 0.1, 0.9 on the KITTI data set. 
Parameters $\lambda_{col}$, $\lambda_{grad}$ and $\lambda_{disp}$ are the truncate costs. In the occluded area of the image and the blank area in the disparity map, the truncation cost can limit the aggregation cost to a rational range and increase the robustness of the algorithm.
The parameters $\lambda_{col}$, $\lambda_{grad}$ and $\lambda_{disp}$ can be set to 10, 2 and 2. 
%	However, the truncation cost also filter the more precise cost. 
%	If the data accuracy is relatively high, reducing the value of the truncation cost would increase the accuracy.
%	If you want higher accuracy or have better quality data, you can use a smaller truncation cost. 
In \cite{hosni2012fast}, Rhemann et al. proposes the matching cost of color and gradient, but the addition of LiDAR data allows the algorithm to have higher accuracy and robustness.
\begin{align}
	\alpha +\beta+ \gamma = 1
\end{align}
To avoid the final aggregation cost value being too large, the sum of $\alpha$, $\beta$ and $\gamma$ is limited to 1.
Because of adding LiDAR data, this equation has more features. If these parameters are all greater than 0, this method still has some robustness in the area that LiDAR data doesn't exist. But if $\beta$ is set to 0, this method would achieve the best performance. The time is reduced by $10\%$ and the accuracy is also improved to some extend.  
\subsection{Minimize matching cost}
This section is used to determine the parameters of the plane that minimize the matching cost. 
%	An improved method is proposed by this article to get a higher calculation speed and accuracy. 
%If parallel calculation is used on GPU, the calculation time can be further reduced.
The algorithm can be divided into three steps: initialization, disparity propagation, and plane refinement. 
The algorithm performs disparity propagation and plane refinement once every iteration. 
In this article, the algorithm performed three iterations. 
And in the following subsection, some details of the algorithm can be introduced.
\subsubsection{Initialization}
The plane model has been determined in the section of the window model. Then each pixel needs to be assigned a plane, but many plane parameters have not been determined such as $x$, $y$, $d$, $n_{x}$, $n_{y}$, $n_{z}$. So each pixel can be assigned a random unit vector $n_{x}, n_{y}, n_{z}$.
If the window model is the front-parallel windows, the unit vector is $\textbf{0}$.
And if the $p^{'}(x,y)$ has disparity $d^{'}$ in the up-sampling disparity map, the disparity $d^{'}$ would be assigned to $d$. If no such value exists, a random initial value between the maximum and minimum disparity of the whole image would be assigned to $d$. 
\subsubsection{Disparity propagation}
\begin{figure}[!htbp] %%跨栏图为：figure*而单栏图为 figure
	\centering
	\subfloat{ \includegraphics[width=3.5in]{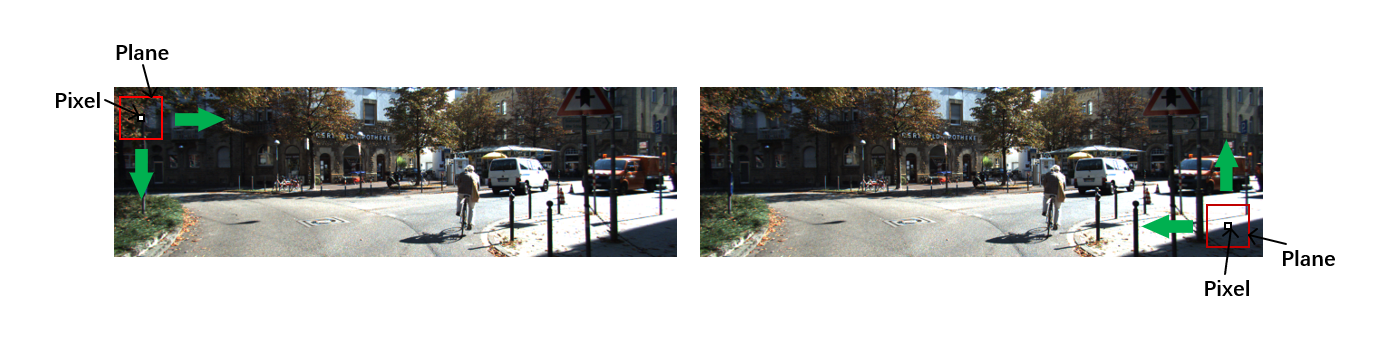} }
	\quad
	\caption{The pixel propagation direction.}
	\label{propagation}
\end{figure}  %%跨栏图为：figure*而单栏图为 figure
Each pixel has been assigned a plane, and then the plane with the least cost of aggregation on each pixel needs to be found.
Because adjacent pixels are likely to have similar planes, the planes with a relatively low aggregation cost can be found by spreading the planes of adjacent pixels. There are many ways to spread, such as from left to right, from top to bottom, from top left to bottom right, and so on. 
%	Referring to the SGM algorithm \cite{hirschmuller2007stereo}, each pixel also finds four directions for propagation. 
The method used in this article is proposed in \cite{bleyer2011patchmatch}. If the number of propagation is odd, the propagation direction from top-left to bottom-right is described as the right part of Fig. \ref{propagation}. If the number of propagation is even, the propagation direction is from the bottom-right to the top-left as the left part of Fig. \ref{propagation}. If $C(p,\xi_{p}) > C(p,\xi_{p})_{right}$ for pixels in the image, $C(p,\xi_{p})=C(p,\xi_{p})_{right}$ need to be done. And then if $C(p,\xi_{p}) > C(p,\xi_{p})_{down}$, $C(p,\xi_{p})=C(p,\xi_{p})_{down}$ need to be executed among adjacent pixels in the odd number of propagation. Pixel $p$ needs to be compared with $p_{up}$ and $p_{left}$ in the process of even-numbered propagation. 

\begin{figure*}[!htb]
	\centering
	\subfloat[]{\includegraphics[width=0.3\textwidth]{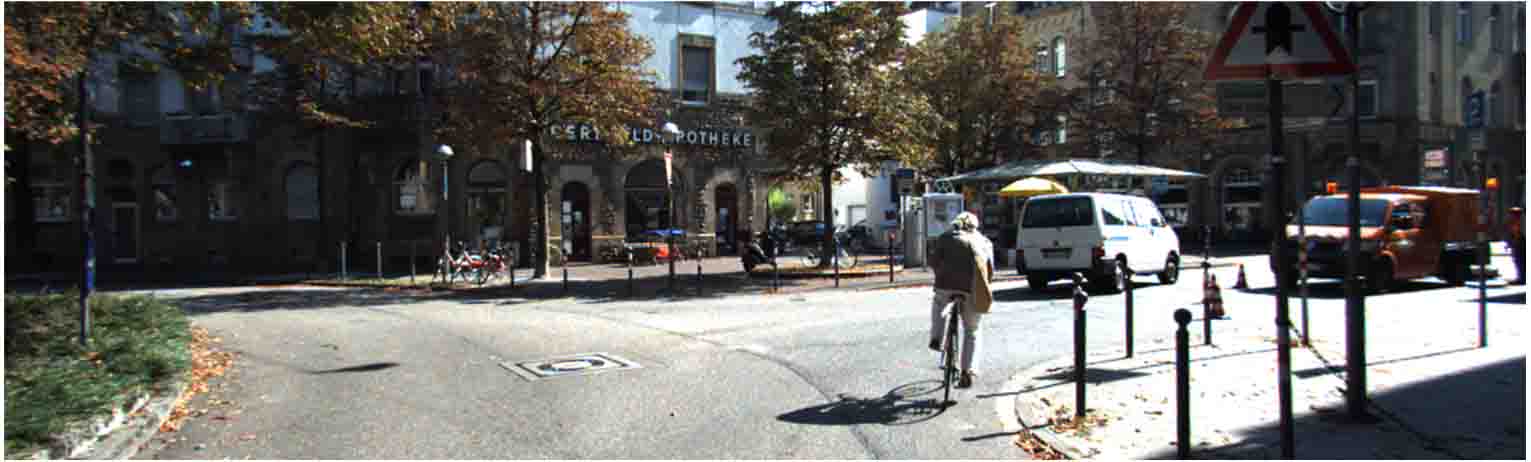}}	
	\subfloat[]{ \includegraphics[width=0.3\textwidth]{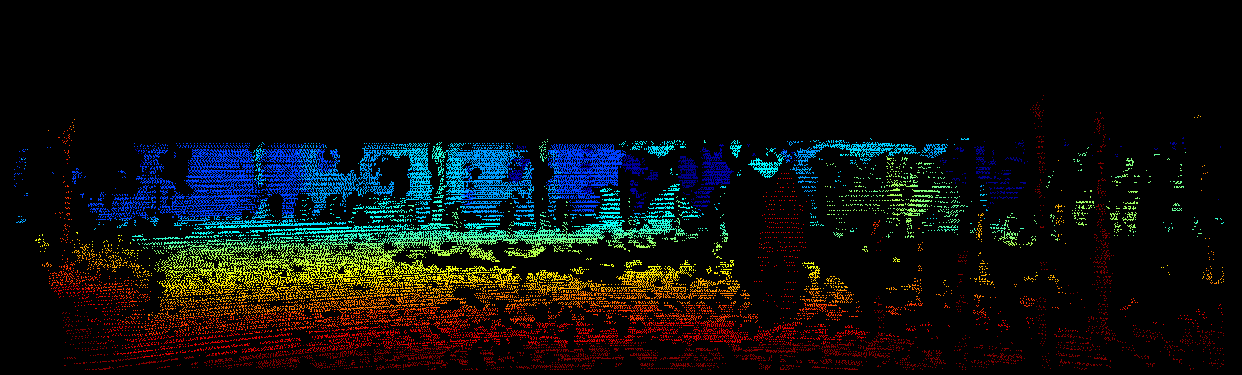} }
	\subfloat[]{ \includegraphics[width=0.3\textwidth]{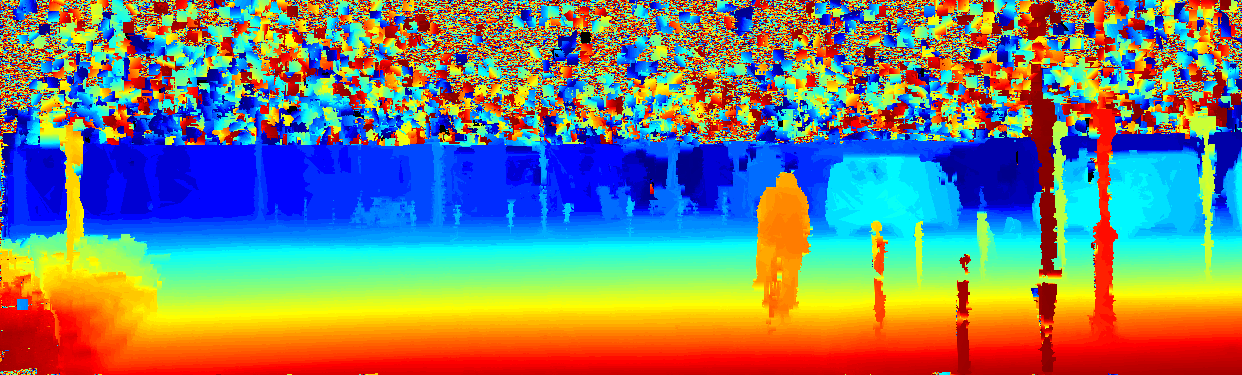} }
	\quad
	\subfloat[]{\includegraphics[width=0.3\textwidth]{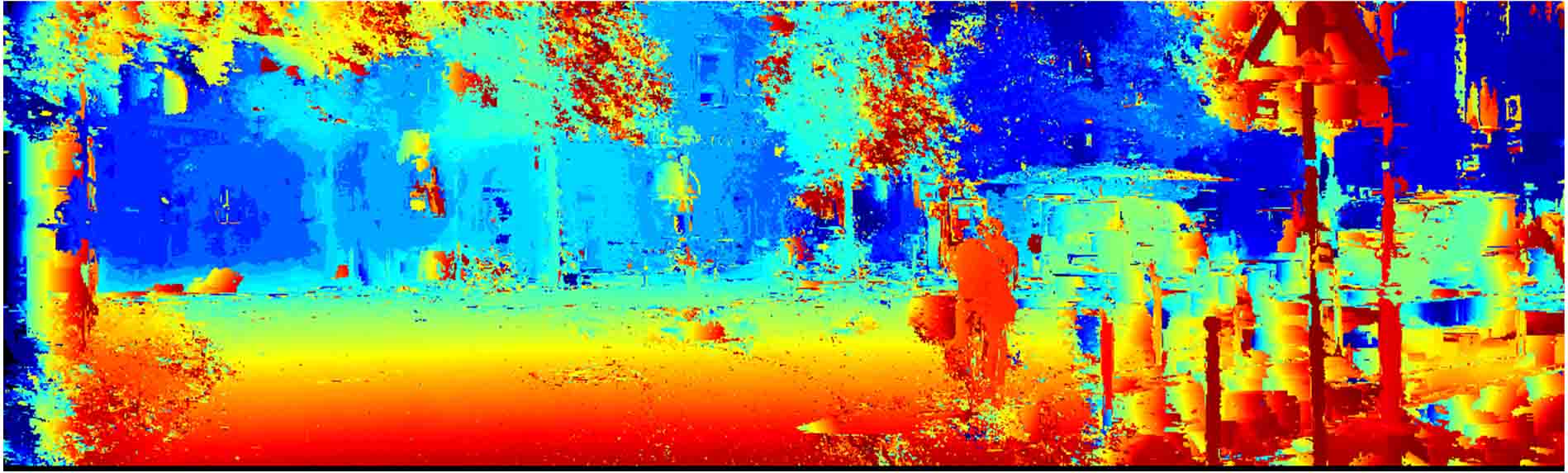}}	
	\subfloat[]{ \includegraphics[width=0.3\textwidth]{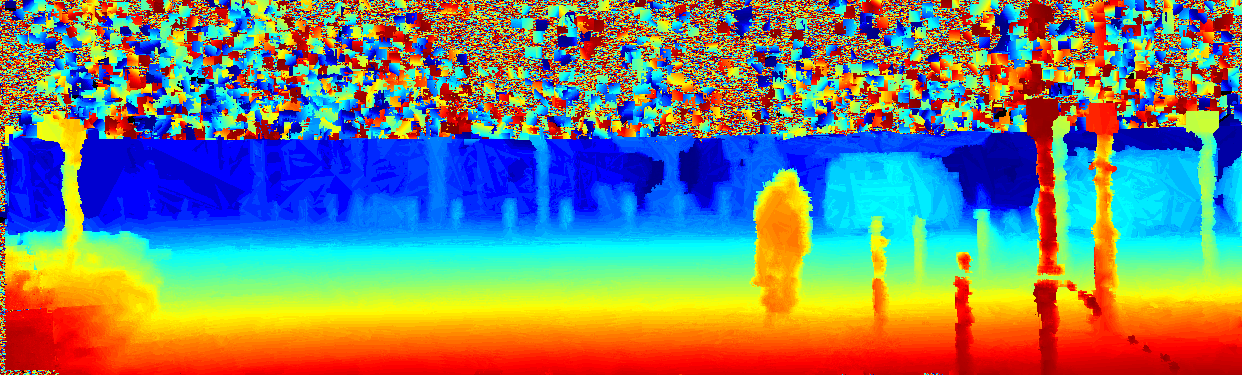} }
	\subfloat[]{ \includegraphics[width=0.3\textwidth]{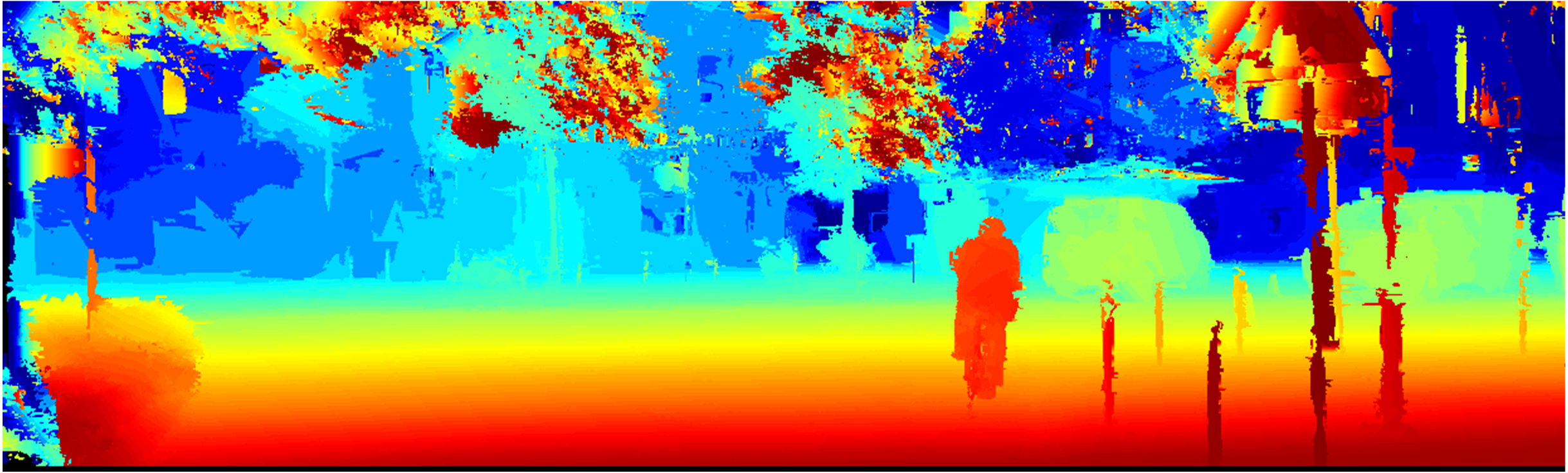} \label{fig5:f}}
	\quad
	\subfloat[]{\includegraphics[width=0.3\textwidth]{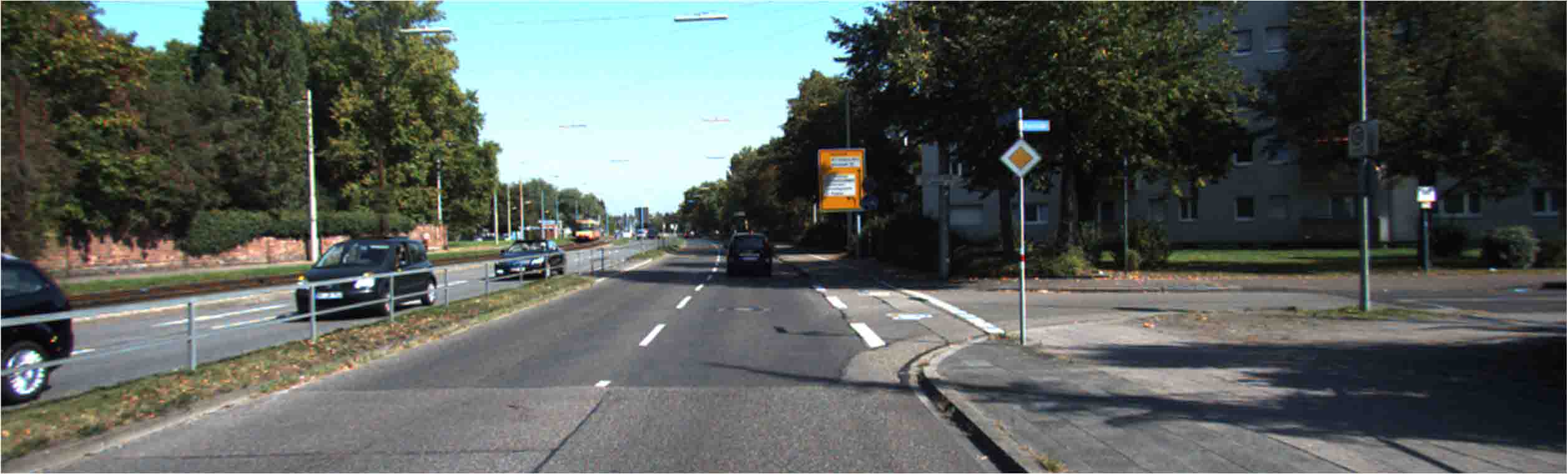}}	
	\subfloat[]{ \includegraphics[width=0.3\textwidth]{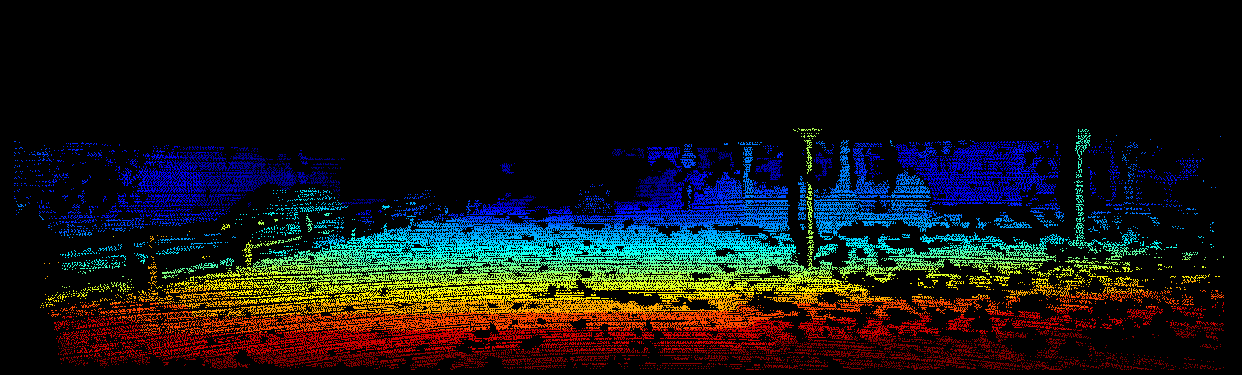} }
	\subfloat[]{ \includegraphics[width=0.3\textwidth]{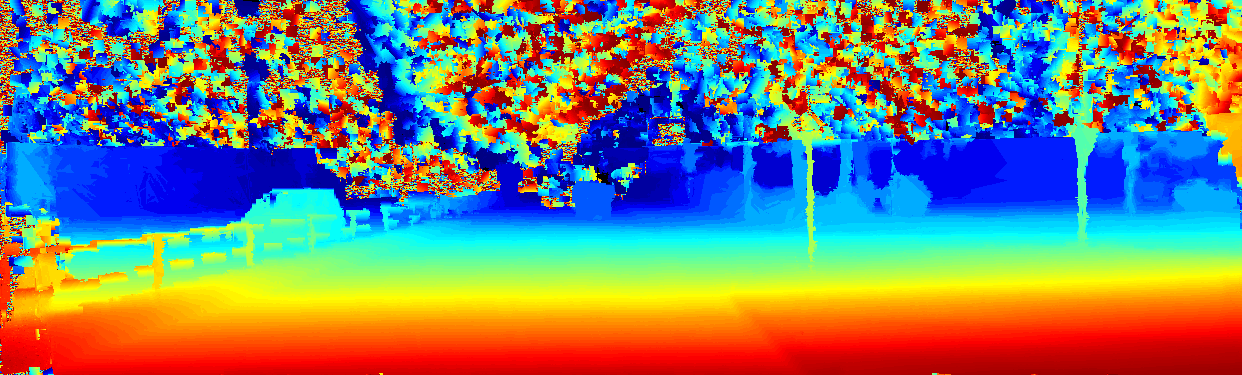} }
	\quad
	\subfloat[]{\includegraphics[width=0.3\textwidth]{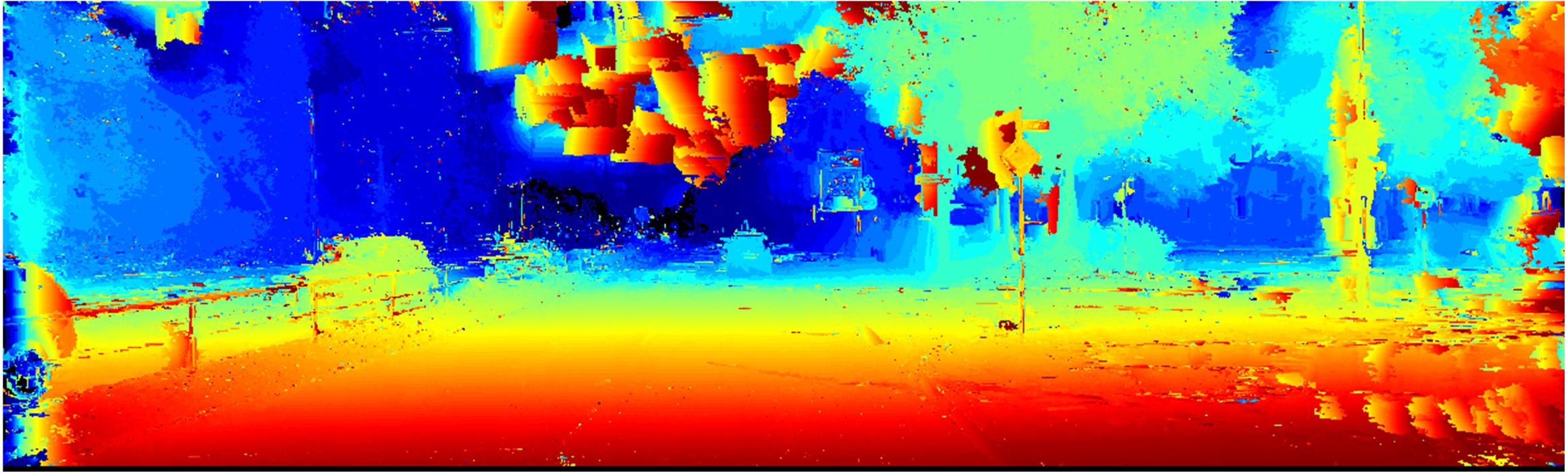}}	
	\subfloat[]{ \includegraphics[width=0.3\textwidth]{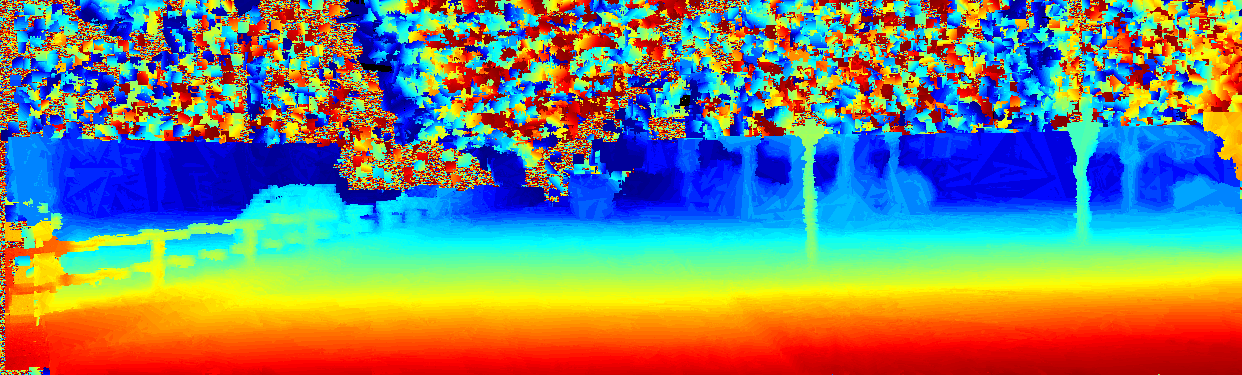} }
	\subfloat[]{ \includegraphics[width=0.3\textwidth]{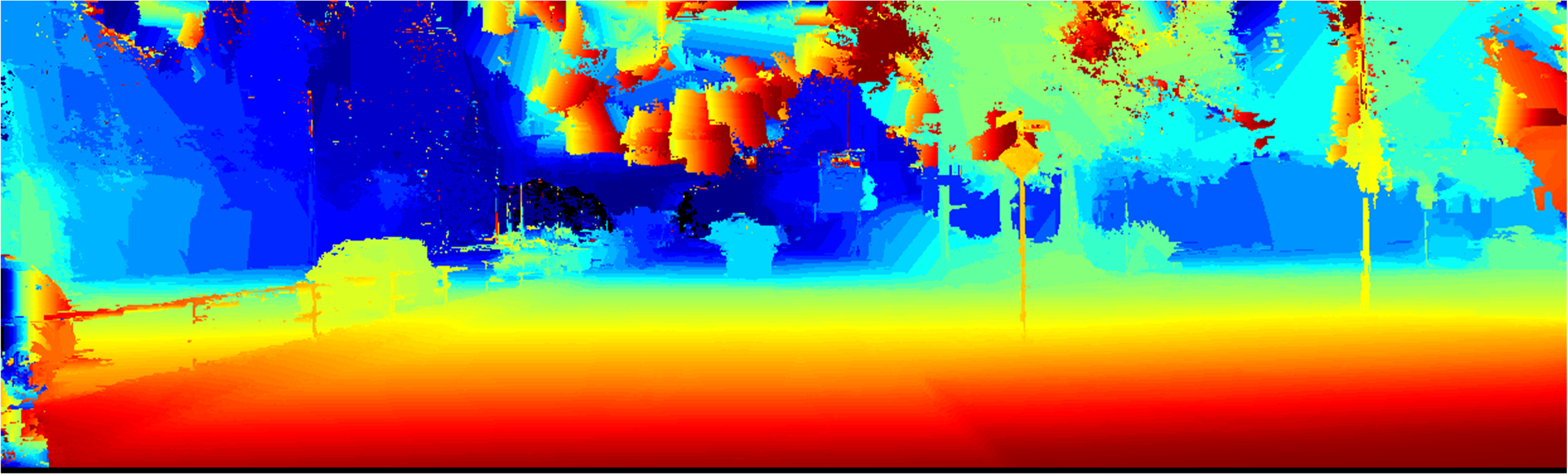} \label{fig5:l}}
	\quad		
	%		\caption{The results on KITTI. (a) and (g) are the left image of stereo images. (b) and (h) are ground truth of the disparity maps. (c) and (i) are results of ours($BI^{*}$).  (d) and (j) are results of The PMS.  (e) and (k)are results of ours($BF^{*}$).  (f) and (l) are results of ours($BI^{*}$), window size of which is 35.}
	\caption{The results on KITTI. (a)(g) the left image of stereo images. (b)(h) the ground truth of depth completion. (c)(i) the results of OURS($BI^{*}$). (d) (j) the results of PMS.  (e)(k) the results of OURS($BF^{*}$). (f)(l) are the results of OURS($BI^{*}$) when $\alpha$, $\beta$, and $\gamma$ are set to $0.1$, $0.1$, and $0.8$ and window size of which is from 10 to 35. And then the robustness of LiDAR data missing areas is enhanced.}
	\label{fig5}
\end{figure*}

\subsubsection{Plane refinement}
In order to further reduce the aggregation cost of Eq.\ref{cost}, the plane needs to be refined. The refined plane is between the two propagation processes, and it provides the new plane with a lower aggregation cost for the propagation process if possible. This is also the most time-consuming step of the algorithm. We make use of parallel processing method to reduce calculation time of this subsection. 
%	We assume that when the pixel is in the middle of the two scan lines, its disparity will also be distributed between the disparity of the two scan lines.
In this process, the plane $\xi$ need to be represented a point $(x,y,d)$ and a normal vector $(n_{x},n_{y},n_{z})$. Two parameters $\Delta_{d}^{MAX}$ and $\Delta_{n}^{MAX}$ are defined as the maximum allowed range of the disparity $d$ and $n$. If the pixel is located between two LiDAR scan lines, $\Delta_{d}^{MAX}=\lvert\Pi(x,y_{up})-\Pi(x,y_{down})\rvert$. If not, $\Delta_{d}^{MAX}$ would be the maximum disparity of the image. Similarly, $\Delta_{n}^{MAX}$ is also the maximum allowable range of the vector. 
$\Delta_{d}$ is randomly selected from $(-\Delta_{d}^{MAX},\Delta_{d}^{MAX})$ to calculate $d^{'}=d+\Delta_{d}$.
$\vec{ \Delta_{n} }$ is also a random value in the $(-\Delta_{n}^{MAX},\Delta_{n}^{MAX})$ and $\vec{n^{'}}=unit(\vec{n}+\vec{\Delta_{n}})$, where unit() is unitized.
Now all parameters of a new plane $\xi^{'}$ are determined.
If $C(p,\xi_{p}) > C(p,\xi_{p}^{'})$, $\xi_{p}^{'}$ can be used instead of $\xi_{p}$. 
This is an iterative process. For the first time, $\Delta_{d}^{MAX}$ is set as $\Delta_{d}^{MAX} / 2$ and $\Delta_{n}^{MAX} = 1$. 
$\Delta_{d}^{MAX}$ is updated as$\Delta_{d}^{MAX} / 2$ and $\Delta_{n}^{MAX}=\Delta_{n}^{MAX} /2$ every iteration. This can reduce our search range. If $\Delta_{d}^{MAX}<0.1$, refining the plane $\xi_{p}$ is aborted.
In the first iteration, the disparity changes may be large. In subsequent iterations, the disparity changes may gradually decrease. 
%	If the planes differ greatly, they may be adjusted in the first iteration. The plane can gradually be refined in subsequent iterations.
After refining the plane was finished, the plane can be able to restart disparity propagation again. 
%	In our experiment, a total of three operations of spreading and refining the plane were carried out.

\section{EXPERIMENT AND ANALYSIS}
\subsection{Experiment setup and setting}
\begin{figure}
	\centering
	\subfloat{ \includegraphics[width=3.1in]{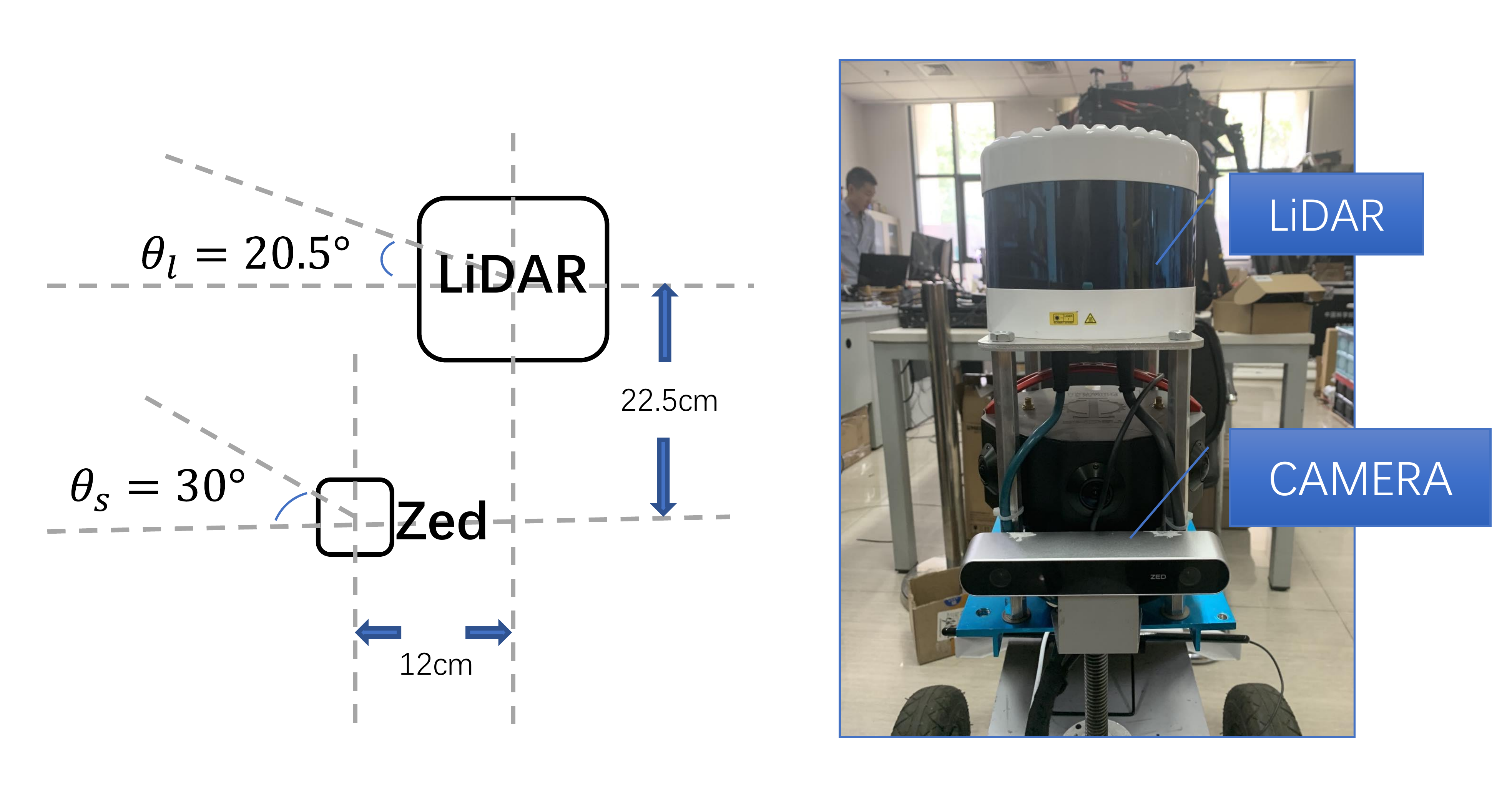} }	
	\caption{Experiment device. The stereo camera captures left and right images and the LiDAR acquires point clouds.}
	\label{setup}
\end{figure}

Our experimental device consists of a stereo camera and 3D LiDAR, as shown in the right part of Fig. \ref{setup}. The left part of Fig. \ref{setup} shows the installation method of the one. Stereo images and LiDAR point clouds can be captured by this device.
%	, but the ground truth of the disparity map is not easy to obtain. 
To better calculate the error, we evaluate the performance of our algorithm on the KITTI depth benchmark \cite{uhrig2017sparsity}.
The data we use comes from the KITTI Raw benchmark \cite{kitti}, which is the raw data under different road environments.
%	The raw LiDAR data and the ground truth of the KITTI stereo benchmark are different. The raw LiDAR data only include one frame. But the ground truth is that a set of consecutive frames about 11 frames are registered using ICP.\cite{kitti} And all vehicles are annotated by CAD models and all outliers mentioned in section 3 are removed in LiDAR frames.\cite{menze2015object}
The experimental device of KITTI consisted of Velodyne HDL-64E and two FLIR color cameras.
The KITTI depth benchmark contains 46k ground truth. We use 1k ground truth containing different scenes for testing, because the test set of the KITTI depth benchmark does not provide the stereo images and the ground truth. 
We also note that we test the area covered by the raw LiDAR data.
%	, because of the limitation of our method. 
%	The loss of LiDAR data is mainly due to the detection range of LiDAR.
%	We use the KITTI stereo benchmark training dataset for testing. It has a total of 140 pairs of stereo images covering 28 scenes, the LiDAR data of which can be found in the KITTI raw benchmark.
%	Because the KITTI2015 stereo test set does not include raw LiDAR data. 
%	We use the KITTI2015 stereo training set provides 130 stereo images that include the corresponding LiDAR point cloud. 
To ensure the correctness and fairness of the evaluation process, the evaluation development kit is used to evaluate, which is provided by the KITTI depth benchmark. 
The configurations of our computer are i7-8700 and 8G RAM.
%	Moreover, KITTI provides a lot of raw data, a subset of which is used to test. 
For readers to better understand the characteristics of our methods, we compare them in five different ways. The results are shown in Fig. \ref{fig6} and TABLE. I- IV. In these tables, "$>2px$" means that errors greater than 2 pixels are counted, like also "$>3px$" and "$>5px$". 
%	The three columns "D1-fg", "D1-bg", and "D1-all" mean that percentage of stereo disparity outliers averaged only over foreground regions, background regions, and all ground truth areas in the first frame. The KITTI2015 stereo benchmark considers a pixel to be correctly estimated if the disparity or flow end-point error is less than $3$ pixels or $5\%$ of its real value\cite{menze2015object}. 
OURS($BI^{*}$) and OURS($BF^{*}$) represent results of combining stereo images with $BI^{*}$ and $BF^{*}$. 
RMSE, MAE, iRMSE and iMAE \cite{wong2021unsupervised} mean root mean squared error, mean absolute error, and ones of the inverse depth. 
RMSE and MAE indicate the performance of the algorithm at distant location.
iRMSE and iMAE increase the weights of small depth. So they can represent the performance of the algorithm at close location.
We get the experience data when $\alpha$, $\beta$, and $\gamma$ are set to $0.1$, $0$, and $0.9$.
\subsection{Impact of window size}
\begin{figure}[!htbp] %%跨栏图为：figure*而单栏图为 figure
	\centering
	\subfloat[The influence of different window sizes on error with slanted support window]{ \includegraphics[width=3.5in]{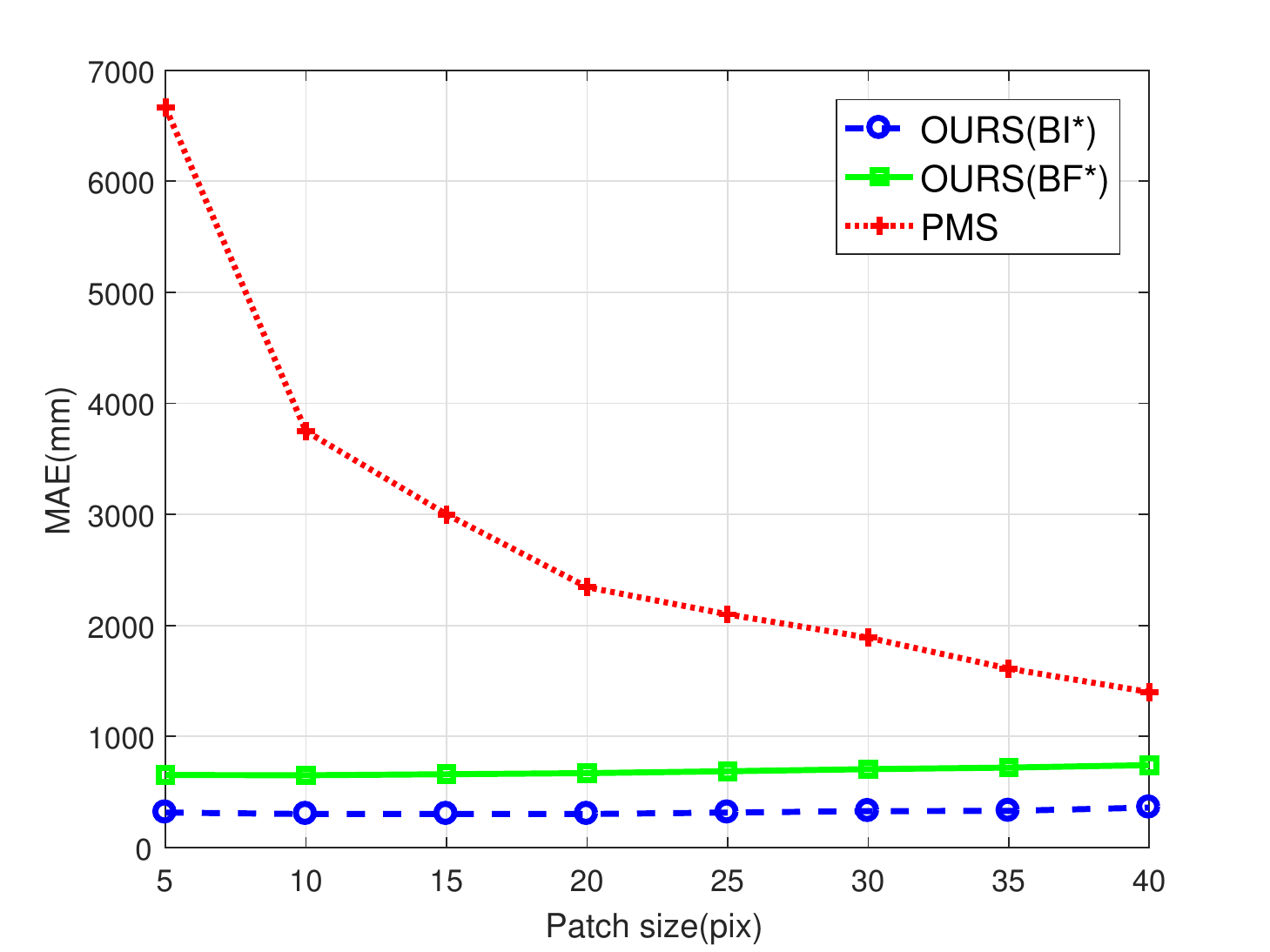} \label{fig6:a}}
	\quad
	\subfloat[The influence of different window sizes on time with slanted support window]{ \includegraphics[width=3.5in]{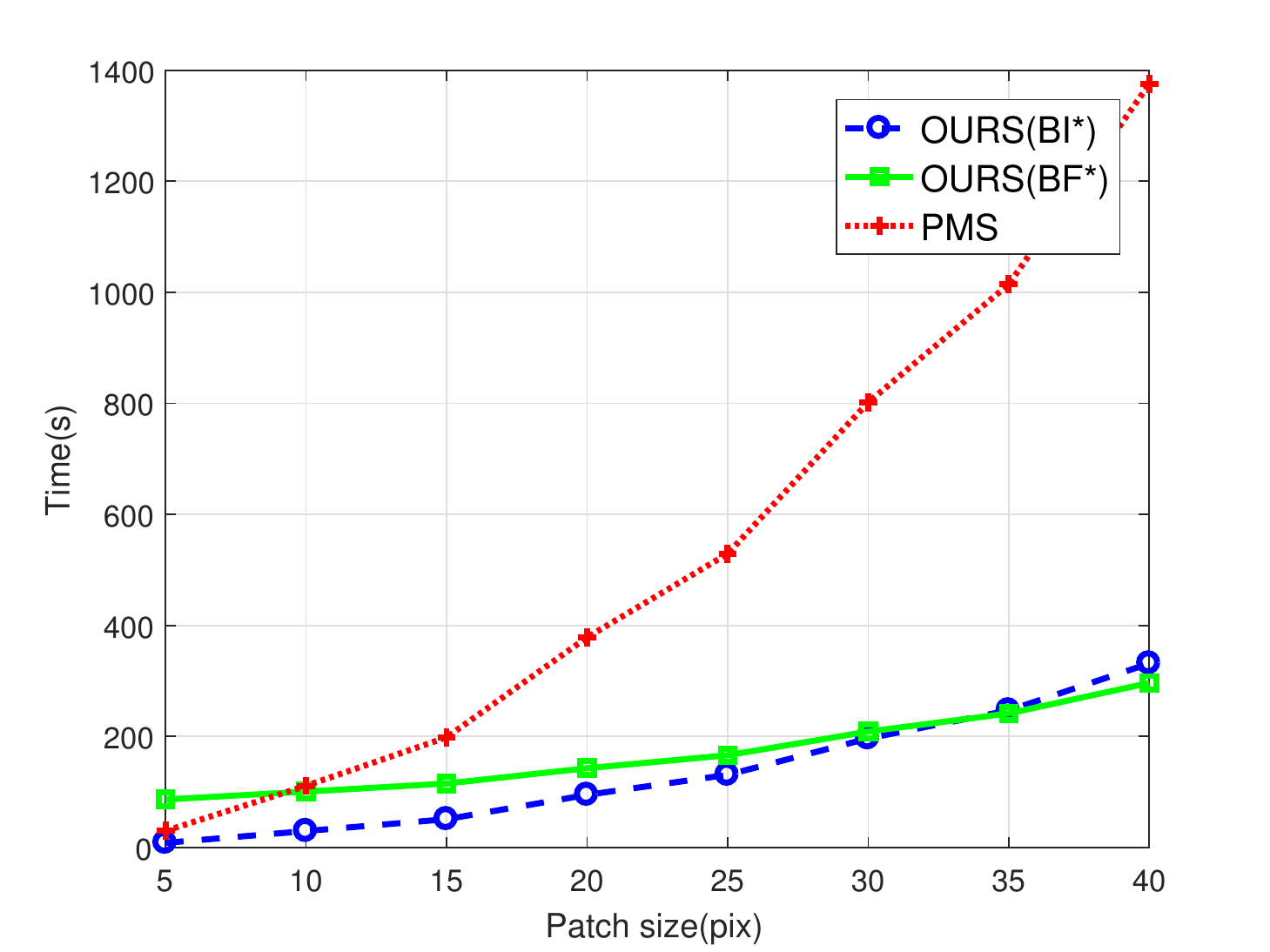} \label{fig6:b}}
	\quad
	\caption{Performance of our methods and PMS in different window sizes.}
	\label{fig6}
\end{figure}

The patch size is an important parameter, which has a great influence on accuracy and time. Fig. \ref{fig6} introduces the relationship between window size, MAE, and time. 
Due to the existence of outliers, the use of a single pixel for matching would be affected by the outliers. But using all pixels in the window for matching would eliminate the influence of outliers to some extent.
The window size of PMS is 40 that is the minimum error as illustrated in Fig. \ref{fig6:a}. 
In subsequent experiments, the patch size of PMS is set to 40.
PMS is similar to the traditional stereo matching algorithm, which makes the algorithm more robust when increasing the window size appropriately.
Compared with PMS, the error of our methods is not sensitive to window size. The addition of LiDAR data improves the robustness of the algorithm.
Our methods have the smallest error as shown in fig. \ref{fig6:a} when the window size is approximately 10. 
%	But a larger window can increase the robustness of the LiDAR missing area as shown in \ref{fig5:f} and \ref{fig5:l}.
%	Because, if the window size is too large, the pixels at the edge of the plane and the pixels in the middle may not be in the same plane, causing a relatively large error.
In addition, increasing the window size would increase the time for all methods as shown in Fig. \ref{fig6:b}, because a larger window size requires more calculations. 
The calculation speed of OURS($BI^{*}$) is 3.55 times that of PMS. 	
At the beginning of the curve of OURS($BF^{*}$), $BF^{*}$ occupies most of the time. 
In addition, because the up-sampled disparity map of $BF^{*}$ is denser than that of $BI^{*}$ as shown in Fig.\ref{BF}, the calculation time of OURS($BF^{*}$) is shorter than that of OURS($BI^{*}$) to some extent at the ending of the curve of OURS($BF^{*}$). 
  %%跨栏图为：figure*而单栏图为 figure
\subsection{Upsampling experiment on LiDAR disparity map}
\begin{table}[!htbp]
	\centering
	\caption{Up-sampling Experimental}
	\begin{tabular}{llllll}
		\hline
		Method    & \textgreater{}2px & \textgreater{}3px & \textgreater{}5px   \\
		\hline
		$BI^{*}$  & \bm{$4.52$\%}    & \bm{$3.73$}\%    & 3.27\%     \\
		$BF^{*}$     &16.13\%     & 5.74\%   & \bm{$2.87$}\%     \\ 
		\hline
		\label{upsample}
	\end{tabular}	
\end{table}
$BF^{*}$ is a method based on bilateral filtering\cite{tomasi1998bilateral}, which makes the distribution range of disparity smaller. 
This may be not conducive to spreading the correct disparity, because it may magnify some errors when the truth data is far from the mean value. 
%	There is a big gap between the generated up-sampling data and truth data.
However, $BI^{*}$ is an interpolation method, which can fill the gap between two scan lines very well when the LiDAR beam is relatively dense.
Because the surface of the object is divided into many different scan lines, the idea of differential geometry is used to understand the relationship between the surface of the object and the LiDAR scanning line.

For the above reasons, we can find that the accuracy of $BI^{*}$ is higher than that of $BF^{*}$ when the error is relatively small from TABLE \ref{upsample}. The pixel distribution of $BF^{*}$ is more concentrated.  A large part of the error of $BF^{*}$ is distributed in "$>2px$", but the error in "$>5px$" is relatively small.
\begin{table*}[!hbtp]
	\centering
	\caption{Experimental results of different methods}
	\begin{tabular}{lllllll}
		\hline
		Method     &Modality  &RMSE(mm) &MAE(mm) &iRMSE(1/km )  &iMAE(1/km)  &Time(s)\\
		\hline
		%		$BF^{*}$ \cite{premebida2016high} & 4.10\%  &\bm{$5.10$}\% & 3.95\% \\
		%SGM\cite{hirschmuller2007stereo}          & 6.74 & 8.09 & 6.50 \\
		%PMS\cite{bleyer2011patchmatch}			&Stereo        &4776 &1611   &18.15   &6.10 &1375  \\
		%PSM-NET \cite{chang2018pyramid}  &Stereo  & 885   &332  & 1.65 &1.00 &0.41 \\
		%		Maddern \textit{et al.}\cite{maddern2016real} &Stereo+LiDAR &6.08 & 13.56 &5.56 &  &0.03  \\
		Park \textit{et al.} \cite{park2018high} &Stereo+LiDAR  &2021  &500  &3.39   &1.39 &0.04   \\
		LiStereo(self-supervised) \cite{zhang2020listereo}    &Stereo+LiDAR  &1278 &326  &3.83   &1.32 &-  \\
		LiStereo(supervised) \cite{zhang2020listereo}    &Stereo+LiDAR  &834 &284  &2.20   &1.10 &-  \\
		CCVN \cite{wang20193d}       &Stereo+LiDAR  &750  &253  &\textbf{1.40}  &0.81 &1.01 \\
		VPN \cite{choe2021volumetric}         &Stereo+LiDAR  &\textbf{640}  &\textbf{206}  &1.88   &1.00 &-  \\
		OURS($BI^{*}$)& Stereo+LiDAR  & 1282   &302     &1.65 &\textbf{0.80} &6.01 \\
		\hline
		\label{table2}
	\end{tabular}	
\end{table*}
\subsection{Slanted support window experiment}
\begin{table}[!htbp]
	\centering
	\caption{Slanted support window experiment}
	
	\begin{tabular}{llllll}
		\hline
		Method    &RMSE &MAE &iRMSE  &iMAE \\
		\hline
		PMS       &4776 &1611   &18.15   &6.10   \\
		OURS($BF^{*}$) & 1787   &651     &3.48 &2.43 \\
		OURS($BI{*}$)   & \textbf{1282}   &\textbf{302 }    &\textbf{1.65} &\textbf{0.80} \\
		\hline
		\label{table1}
	\end{tabular}	
\end{table}
We use different up-sampling methods to estimate the depth, such as $BI^{*}$ and $BF^{*}$ of \cite{premebida2016high}. 
The TABLE. \ref{table1} compares the performance of our fusion methods and PMS algorithm. After adding LiDAR data, the algorithm performance has been greatly improved. 
%	But Compared with $BF^{*}$, the accuracy of the OURS ($BF^{*}$) method is worse at "$>2px$" and "$>3px$".
%	The filtering method will make the distribution range of the disparity smaller, which is not conducive to the spread of the disparity.
And OURS($BI{*}$) is better than OURS($BF^{*}$), because the evenly distributed disparity is more conducive to spreading.
%	LiDAR points and points generated by $BF^{*}$ are usually not in the same disparity plane. The wrong disparity map is used for the disparity propagation process, which will magnify the error.
% 	Since the number of LiDAR points is very small, they are usually regarded as outliers of the disparity plane.
%	In addition, the outlier of LiDAR points existing at the edge of the foreground will also affect the calculation of the matching cost as shown in Fig.\ref{fig1:b}.
It has a great influence on searching for the best disparity plane.
Therefore, the experimental results show that the performance of OURS($BI^{*}$) is better than the other two methods. Fig. \ref{fig5} shows the results of three algorithms. And Fig. \ref{fig5:f} and Fig. \ref{fig5:l} indicate that adding gradient information and increasing window size can enhance the accurate and robustness in the area of lidar data missing.
%But OURS($BI{*}$) and OURS($BF{*}$)'s performance are decreased when the lidar data is missing.  
%	We get the experience images when $\alpha$, $\beta$, and $\gamma$ are set to $0.1$, $0.1$, and $0.8$ and the patch size is set to $10$. 
%\begin{figure*}[!htbp] %%跨栏图为：figure*而单栏图为 figure
%	\centering
%	\subfloat[]{\includegraphics[width=0.3\textwidth]{pic/060/picture13.jpg}}	
%	\subfloat[]{ \includegraphics[width=0.3\textwidth]{pic/060/picture14.jpg} }
%	\subfloat[]{ \includegraphics[width=0.3\textwidth]{pic/060/picture15.png} }
%	\quad
%	\subfloat[]{\includegraphics[width=0.3\textwidth]{pic/060/picture16.png}}	
%	\subfloat[]{ \includegraphics[width=0.3\textwidth]{pic/060/picture17.png} }
%	\subfloat[]{ \includegraphics[width=0.3\textwidth]{pic/060/picture18.png} }
%	\quad
%%	\caption{Results of different algorithms. (a) is the left image of stereo images. (b) is ground truth of the disparity map. (c) is the results of Maddern et al.\cite{maddern2016real}. (d) is the results of SGM\cite{hirschmuller2007stereo}. (e) is the results of PSM-NET\cite{chang2018pyramid}.  (f) is the results of ours($BI^{*}$). }
%	\caption{Results of different algorithms. (a) the left image of stereo images. (b) ground truth of the disparity map. (c) the results of Maddern et al.\cite{maddern2016real}. (d) the results of SGM\cite{hirschmuller2007stereo}. (e) the results of PSM-NET\cite{chang2018pyramid}. (f) the results of OURS($BI^{*}$). }
%	\label{diferentmethod}
%	\end{figure*}  %%跨栏图为：figure*而单栏图为 figure

\subsection{Front-parallel window experiment}
\begin{table}[!htbp]
	\centering
	\caption{Front-parallel window experiment}
	\begin{tabular}{llllll}
		\hline
		Method   &RMSE &MAE &iRMSE  &iMAE \\
		\hline
		PMS            &6457 &2345   &24.92   &9.70   \\
		OURS($BF^{*}$)  &1760 &655   &3.58   &2.51   \\
		OURS($BI^{*}$)  &\textbf{1569} &\textbf{311}   &\textbf{2.45}   &\textbf{0.92 }  \\
		\hline
		\label{frontparallel}
	\end{tabular}	
\end{table}
The performance of the slanted support window is better than that of the front-parallel window.
But the performance improvement of our methods is not as significant as the PMS uses the slanted support window.
%	The performance improvement mainly comes from the reliable propagation of disparity. The contribution of the slanted support window to our method is limited. 
In addition, OURS($BF^{*}$) has a very small gap between using the slanted support window and the front-parallel window.
This is also caused by the up-sampling method of $BF^{*}$ which makes the disparity map smooth.
The results are shown in the TABLE. \ref{frontparallel}.
%	Although the performance of front-parallel windows is not as good as slanted support windows, it is faster.

%	But the edges of objects are often discontinuous surfaces, which will cause performance degradation of slanted support window as shown in "D1-fg" of TABLE.\ref{frontparallel}. 
%	Slanted support window improves the accuracy of PMS and OURS($BI^{*}$), but it cause the accuracy of OURS($BF^{*}$) to decrease as illustrated in TABLE. \ref{table1} and TABLE. \ref{frontparallel}. 
%	Due to the characteristics of the $BF^{*}$ algorithm as discussed above, OURS($BF^{*}$) using front-parallel windows may result in a smaller matching cost.
%	OURS($BF^{*}$) has the best performance in the "D1-fg" of TABLE. \ref{frontparallel}, because $BF^{*}$ has good edge-preserving capabilities.
\subsection{Comparison of different methods}
%	Due to the noise in the image, it is difficult for the traditional stereo matching approach to overcome these noises to obtain a perfect result. In addition, according to the characteristics of the stereo, the depth estimation accuracy of the distant target object will be poor. Although PSM-NET and corresponding deep learning methods show high accuracy, they require a lot of data and the generalization ability needs to be improved. Our experiments show that reliable disparity information can correct errors in the stereo matching process.
%	We compare the accuracy of ours with other methods such as PSM-NET\cite{chang2018pyramid}, Maddern et al.\cite{maddern2016real}, and SGM\cite{hirschmuller2007stereo}. Due to the fusion of the advantages of multiple sensors, the performance of our method is greatly improved compared to the single-sensor solution such as SGM and is familiar with the deep learning method PSM-NET. 
%	Fig. \ref{diferentmethod} shows the results of different algorithms. 
%	In order to ensure the accuracy of the data, the test data of other algorithms are obtained from papers or benchmark websites. 
We compare our method with different methods. At present, most scholars use deep learning methods to study the fusion of lidar and stereo. On the one hand, with its powerful performance, supervised learning methods can discover a lot of information from the ground truth to train their network. But in the field of depth estimation, obtaining ground truth is very expensive and difficult. Compared with deep learning methods, our method doesn't limited by the data sets. On the other hand, our method is better than the latest unsupervised learning method \cite{zhang2020listereo} in accuracy. If some scenes do not have the ground truth provided by the dataset to train the neural network, our method would be a good choice. 
%If the lidar data and images are accurate, our method is basically not limited by the scene.
%Because of the use of windows for propagation, the RMSE and iRMSE may be relatively large.
In addition, our method has achieved good results on the inverse depth in the TABLE \ref{table2}. This indicates that our method has better performance in close locations. 
Although our method has made great progress in time relative to PMS, it is still difficult to achieve real-time performance.

\subsection{Experiments on our equipment}
\begin{figure}[!htbp] %%跨栏图为：figure*而单栏图为 figure
	\centering
	\subfloat[]{\includegraphics[width=0.3\textwidth]{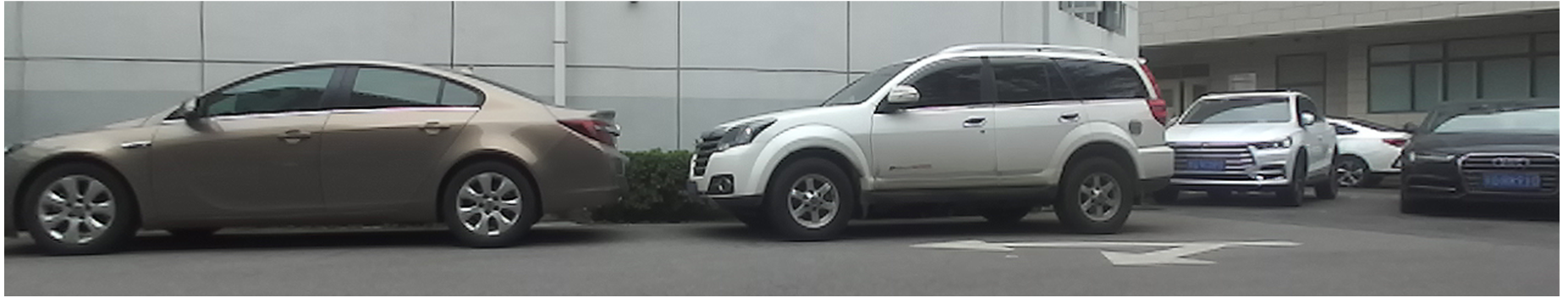}}	
	\quad
	\subfloat[]{ \includegraphics[width=0.3\textwidth]{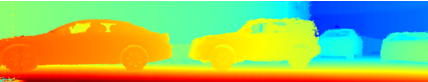} }
	\quad
	%		\subfloat[]{ \includegraphics[width=0.3\textwidth]{pic/test1/图片15.png} }
	\subfloat[]{\includegraphics[width=0.3\textwidth]{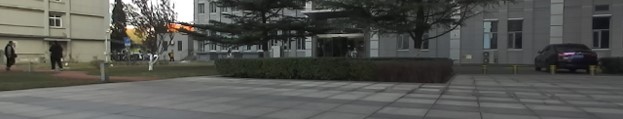}}	
	\quad
	\subfloat[]{ \includegraphics[width=0.3\textwidth]{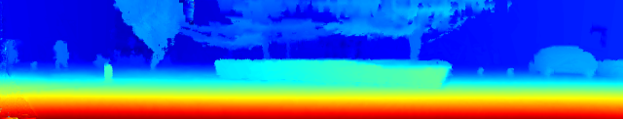} }
	\quad
	%		\subfloat[]{ \includegraphics[width=0.3\textwidth]{pic/test2/picture18.png} }
	
	\caption{The results on our equipment. (a)(c) the left images. (b)(d) the depth map.}
	\label{oursetup}
\end{figure}

This subsection introduces some details in the experiment process.
Our experimental device consisted of Velodyne VLS-128 and ZED stereo camera. We use the autoware autonomous driving software \cite{kato2018autoware} to calculate the positional relationship between the stereo camera and the LiDAR so that the LiDAR points can be transferred to the camera coordinate system. For the camera and LiDAR fusion system to operate normally, the two sensors must collect data at the same time. So timestamps are used to mark LiDAR frames and stereo images. The Velodyne VLS-128 has 128 LiDAR scan lines, but the Velodyne HDL-64E of the KITTI data set has only 64 scan lines. High-performance LiDAR can provide more details for the depth map as shown in Fig.\ref{oursetup}.
%	They are the PMS method, the PMS using the disparity post-processing, the improved method using the linear interpolation, that is, the improved method using the linear interpolation and the disparity post-processing.
%	The last method is to use the up-sampling disparity map proposed in \cite{premebida2016high}. The final calculation result is in TABLE. \ref{table}. We can find that our method is more accurate than the original PMS. On the other hand, after the algorithm uses parallax post-processing, the accuracy will deteriorate. The parallax post-processing will remove part of the original parallax. If the accuracy is not high, the parallax post-processing may improve a certain accuracy like PMS(PP). In the last item of the table, we use bilateral filtering to upsample the disparity map. Bilateral filtering will distort LiDAR point cloud data. The parallax ratio of the two pixels greater than the reference is relatively large. Due to the excellent edge extraction capability of bilateral filtering, the proportion of parallax that is greater than 5 pixels of the benchmark drops quickly. 
  %%跨栏图为：figure*而单栏图为 figure	
\section{CONCLUSION}
Multi-sensor fusion can combine the strengths of various sensors because each sensor has unique characteristics. In detail, LiDAR data is added to the PMS algorithm. The LiDAR depth information and RGB image are fused using the propagation method. Our algorithm greatly improves the accuracy and calculation speed of the PMS algorithm under complex conditions.
The method that we propose is mainly applicable to complex scenes or the object, which is more difficult to estimate the depth. Compared with using the depth camera to calculate a depth map, our method is more suitable for working in large and complex scenes.
Although our method has many several advantages, it also has some disadvantages that need to be improved. For example, it is more dependent on LiDAR data. In the future, we will further study depth estimation methods that are not sensitive to LiDAR data.

\bibliographystyle{IEEEtran}
\bibliography{jsen}

\begin{thebibliography}{10}
\providecommand{\url}[1]{#1}
\csname url@samestyle\endcsname
\providecommand{\newblock}{\relax}
\providecommand{\bibinfo}[2]{#2}
\providecommand{\BIBentrySTDinterwordspacing}{\spaceskip=0pt\relax}
\providecommand{\BIBentryALTinterwordstretchfactor}{4}
\providecommand{\BIBentryALTinterwordspacing}{\spaceskip=\fontdimen2\font plus
\BIBentryALTinterwordstretchfactor\fontdimen3\font minus
  \fontdimen4\font\relax}
\providecommand{\BIBforeignlanguage}[2]{{%
\expandafter\ifx\csname l@#1\endcsname\relax
\typeout{** WARNING: IEEEtran.bst: No hyphenation pattern has been}%
\typeout{** loaded for the language `#1'. Using the pattern for}%
\typeout{** the default language instead.}%
\else
\language=\csname l@#1\endcsname
\fi
#2}}
\providecommand{\BIBdecl}{\relax}
\BIBdecl

\bibitem{kinect}
Z.~Zhang, ``Microsoft kinect sensor and its effect,'' \emph{IEEE multimedia},
  vol.~19, no.~2, pp. 4--10, 2012.

\bibitem{keselman2017intel}
L.~Keselman, J.~Iselin~Woodfill, A.~Grunnet-Jepsen, and A.~Bhowmik, ``Intel
  realsense stereoscopic depth cameras,'' in \emph{Proceedings of the IEEE
  Conference on Computer Vision and Pattern Recognition Workshops}, 2017, pp.
  1--10.

\bibitem{premebida2016high}
C.~Premebida, L.~Garrote, A.~Asvadi, A.~P. Ribeiro, and U.~Nunes,
  ``High-resolution lidar-based depth mapping using bilateral filter,'' in
  \emph{2016 IEEE 19th international conference on intelligent transportation
  systems (ITSC)}.\hskip 1em plus 0.5em minus 0.4em\relax IEEE, 2016, pp.
  2469--2474.

\bibitem{maddern2016real}
W.~Maddern and P.~Newman, ``Real-time probabilistic fusion of sparse 3d lidar
  and dense stereo,'' in \emph{2016 IEEE/RSJ International Conference on
  Intelligent Robots and Systems (IROS)}.\hskip 1em plus 0.5em minus
  0.4em\relax IEEE, 2016, pp. 2181--2188.

\bibitem{shivakumar2019real}
S.~S. Shivakumar, K.~Mohta, B.~Pfrommer, V.~Kumar, and C.~J. Taylor, ``Real
  time dense depth estimation by fusing stereo with sparse depth
  measurements,'' in \emph{2019 International Conference on Robotics and
  Automation (ICRA)}.\hskip 1em plus 0.5em minus 0.4em\relax IEEE, 2019, pp.
  6482--6488.

\bibitem{wang20193d}
T.-H. Wang, H.-N. Hu, C.~H. Lin, Y.-H. Tsai, W.-C. Chiu, and M.~Sun, ``3d lidar
  and stereo fusion using stereo matching network with conditional cost volume
  normalization,'' in \emph{2019 IEEE/RSJ International Conference on
  Intelligent Robots and Systems (IROS)}.\hskip 1em plus 0.5em minus
  0.4em\relax IEEE, 2019, pp. 5895--5902.

\bibitem{choe2021volumetric}
J.~Choe, K.~Joo, T.~Imtiaz, and I.~S. Kweon, ``Volumetric propagation network:
  Stereo-lidar fusion for long-range depth estimation,'' \emph{IEEE Robotics
  and Automation Letters}, vol.~6, no.~3, pp. 4672--4679, 2021.

\bibitem{zhang2020listereo}
J.~Zhang, M.~S. Ramanagopal, R.~Vasudevan, and M.~Johnson-Roberson, ``Listereo:
  Generate dense depth maps from lidar and stereo imagery,'' in \emph{2020 IEEE
  International Conference on Robotics and Automation (ICRA)}.\hskip 1em plus
  0.5em minus 0.4em\relax IEEE, 2020, pp. 7829--7836.

\bibitem{scharstein2003high}
D.~Scharstein and R.~Szeliski, ``High-accuracy stereo depth maps using
  structured light,'' in \emph{2003 IEEE Computer Society Conference on
  Computer Vision and Pattern Recognition, 2003. Proceedings.}, vol.~1.\hskip
  1em plus 0.5em minus 0.4em\relax IEEE, 2003, pp. I--I.

\bibitem{zbontar2016stereo}
J.~Zbontar, Y.~LeCun \emph{et~al.}, ``Stereo matching by training a
  convolutional neural network to compare image patches.'' \emph{J. Mach.
  Learn. Res.}, vol.~17, no.~1, pp. 2287--2318, 2016.

\bibitem{garg2016unsupervised}
R.~Garg, V.~K. Bg, G.~Carneiro, and I.~Reid, ``Unsupervised cnn for single view
  depth estimation: Geometry to the rescue,'' in \emph{European conference on
  computer vision}.\hskip 1em plus 0.5em minus 0.4em\relax Springer, 2016, pp.
  740--756.

\bibitem{godard2017unsupervised}
C.~Godard, O.~Mac~Aodha, and G.~J. Brostow, ``Unsupervised monocular depth
  estimation with left-right consistency,'' in \emph{Proceedings of the IEEE
  conference on computer vision and pattern recognition}, 2017, pp. 270--279.

\bibitem{kitti}
A.~Geiger, P.~Lenz, C.~Stiller, and R.~Urtasun, ``Vision meets robotics: The
  kitti dataset,'' \emph{The International Journal of Robotics Research},
  vol.~32, no.~11, pp. 1231--1237, 2013.

\bibitem{hastie2009overview}
T.~Hastie, R.~Tibshirani, and J.~Friedman, ``Overview of supervised learning,''
  in \emph{The elements of statistical learning}.\hskip 1em plus 0.5em minus
  0.4em\relax Springer, 2009, pp. 9--41.

\bibitem{barlow1989unsupervised}
H.~B. Barlow, ``Unsupervised learning,'' \emph{Neural computation}, vol.~1,
  no.~3, pp. 295--311, 1989.

\bibitem{park2018high}
K.~Park, S.~Kim, and K.~Sohn, ``High-precision depth estimation using
  uncalibrated lidar and stereo fusion,'' \emph{Ieee transactions on
  intelligent transportation systems}, vol.~21, no.~1, pp. 321--335, 2019.

\bibitem{huber2011integrating}
D.~Huber, T.~Kanade \emph{et~al.}, ``Integrating lidar into stereo for fast and
  improved disparity computation,'' in \emph{2011 International Conference on
  3D Imaging, Modeling, Processing, Visualization and Transmission}.\hskip 1em
  plus 0.5em minus 0.4em\relax IEEE, 2011, pp. 405--412.

\bibitem{sander1998density}
J.~Sander, M.~Ester, H.-P. Kriegel, and X.~Xu, ``Density-based clustering in
  spatial databases: The algorithm gdbscan and its applications,'' \emph{Data
  mining and knowledge discovery}, vol.~2, no.~2, pp. 169--194, 1998.

\bibitem{ku2018defense}
J.~Ku, A.~Harakeh, and S.~L. Waslander, ``In defense of classical image
  processing: Fast depth completion on the cpu,'' in \emph{2018 15th Conference
  on Computer and Robot Vision (CRV)}.\hskip 1em plus 0.5em minus 0.4em\relax
  IEEE, 2018, pp. 16--22.

\bibitem{9450836}
H.~Lu, S.~Xu, and S.~Cao, ``Sgtbn: Generating dense depth maps from single-line
  lidar,'' \emph{IEEE Sensors Journal}, vol.~21, no.~17, pp. 19\,091--19\,100,
  2021.

\bibitem{bleyer2011patchmatch}
M.~Bleyer, C.~Rhemann, and C.~Rother, ``Patchmatch stereo-stereo matching with
  slanted support windows.'' in \emph{Bmvc}, vol.~11, 2011, pp. 1--11.

\bibitem{yoon2005locally}
K.-J. Yoon and I.-S. Kweon, ``Locally adaptive support-weight approach for
  visual correspondence search,'' in \emph{2005 IEEE Computer Society
  Conference on Computer Vision and Pattern Recognition (CVPR'05)},
  vol.~2.\hskip 1em plus 0.5em minus 0.4em\relax IEEE, 2005, pp. 924--931.

\bibitem{hosni2012fast}
A.~Hosni, C.~Rhemann, M.~Bleyer, C.~Rother, and M.~Gelautz, ``Fast cost-volume
  filtering for visual correspondence and beyond,'' \emph{IEEE Transactions on
  Pattern Analysis and Machine Intelligence}, vol.~35, no.~2, pp. 504--511,
  2012.

\bibitem{uhrig2017sparsity}
J.~Uhrig, N.~Schneider, L.~Schneider, U.~Franke, T.~Brox, and A.~Geiger,
  ``Sparsity invariant cnns,'' in \emph{2017 international conference on 3D
  Vision (3DV)}.\hskip 1em plus 0.5em minus 0.4em\relax IEEE, 2017, pp. 11--20.

\bibitem{wong2021unsupervised}
A.~Wong and S.~Soatto, ``Unsupervised depth completion with calibrated
  backprojection layers,'' in \emph{Proceedings of the IEEE/CVF International
  Conference on Computer Vision}, 2021, pp. 12\,747--12\,756.

\bibitem{tomasi1998bilateral}
C.~Tomasi and R.~Manduchi, ``Bilateral filtering for gray and color images,''
  in \emph{Sixth international conference on computer vision (IEEE Cat. No.
  98CH36271)}.\hskip 1em plus 0.5em minus 0.4em\relax IEEE, 1998, pp. 839--846.

\bibitem{kato2018autoware}
S.~Kato, S.~Tokunaga, Y.~Maruyama, S.~Maeda, M.~Hirabayashi, Y.~Kitsukawa,
  A.~Monrroy, T.~Ando, Y.~Fujii, and T.~Azumi, ``Autoware on board: Enabling
  autonomous vehicles with embedded systems,'' in \emph{2018 ACM/IEEE 9th
  International Conference on Cyber-Physical Systems (ICCPS)}.\hskip 1em plus
  0.5em minus 0.4em\relax IEEE, 2018, pp. 287--296.

\end{thebibliography}

\begin{IEEEbiography}
	[{\includegraphics[width=1in,height=1.25in,clip,keepaspectratio]{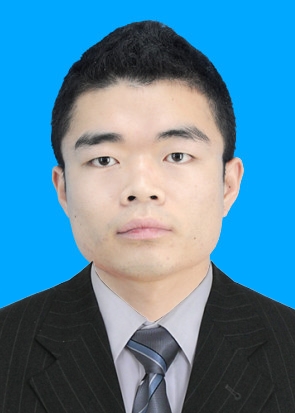}}] {Guangyao Xu}
	received the B.S. degree in computer science and technology from Shandong Jianzhu University, Shandong Province, China, in 2018. He is currently working toward a M.S. degree at the School of Artificial Intelligence, University of Chinese Academy of Sciences, Beijing, China. His research interests include robotics, computer vision, deep learning.
\end{IEEEbiography}
\begin{IEEEbiography}
	[{\includegraphics[width=1in,height=1.25in,clip,keepaspectratio]{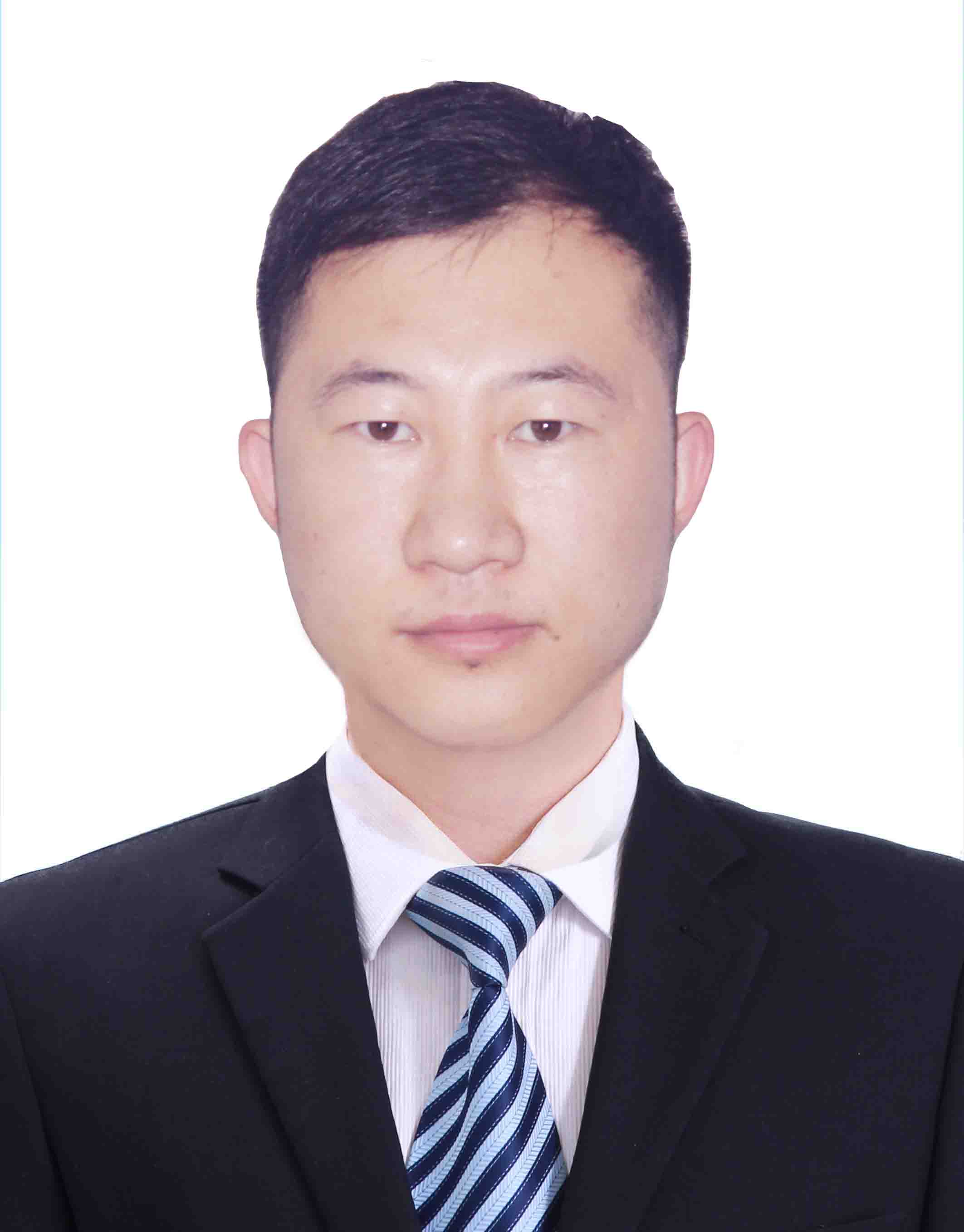}}] {Junfeng Fan}
	received a B.S. degree in mechanical engineering and automation from Beijing Institute of Technology, Beijing, China, in 2014 and a Ph.D. degree in control theory and control engineering from the Institute of Automation, Chinese Academy of Sciences (IACAS), Beijing, China, in 2019. He is currently an Associate Professor of the State Key Laboratory of Management and Control for Complex Systems at IACAS, Beijing, China. His research interests include industrial robotics, robot vision and welding automation.
\end{IEEEbiography}
\begin{IEEEbiography}
	[{\includegraphics[width=1in,height=1.25in,clip,keepaspectratio]{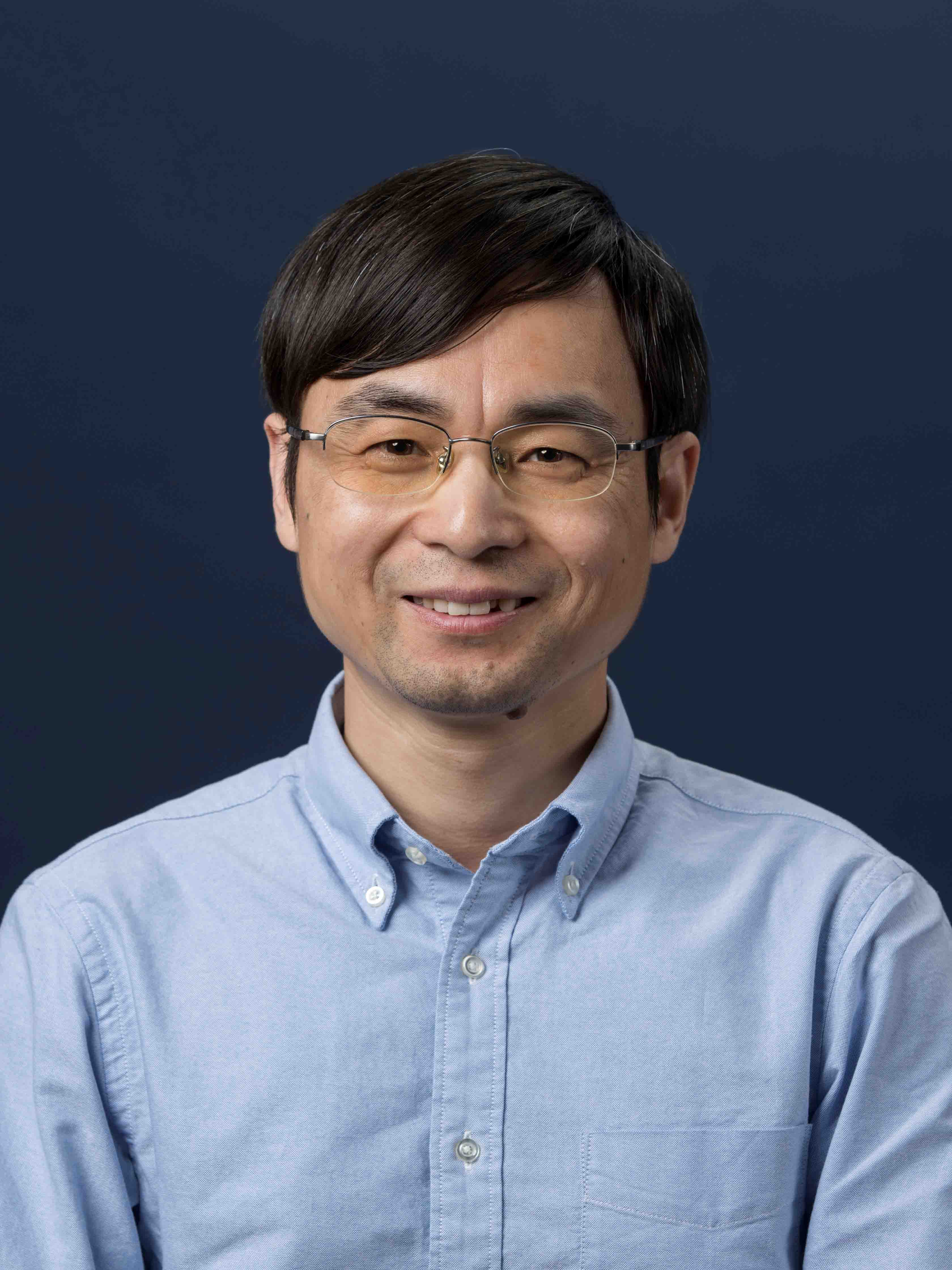}}] {En Li}
	received the B.E. degree in automation from Southwest University of Science and Technology, Mianyang, China,in 2002, and the Ph.D. degree from the Institute of Automation, Chinese Academy of Sciences, Beijing, China, in 2007. He is currently a Professor with the Institute of
	Automation Chinese Academy of Sciences, Beijing. His current research interests include special operation robotic systems, intelligent control, and high redundancy robot technology.
\end{IEEEbiography}
\begin{IEEEbiography}
	[{\includegraphics[width=1in,height=1.25in,clip,keepaspectratio]{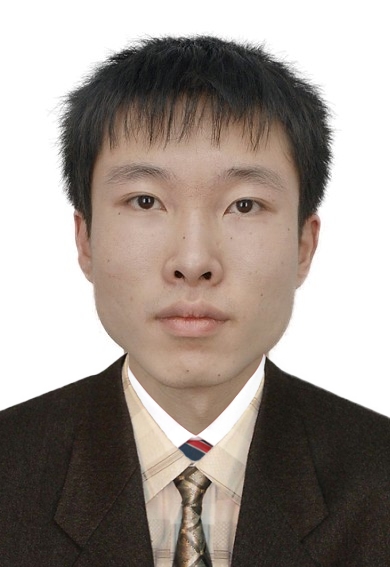}}] {Xiaoyu Long}
	received the B.S. degree in Mechanical Engineering from the University of Nottingham, Ningbo, Zhejiang Province China, in 2017. He is currently working toward a M.S. degree at the School of Artificial Intelligence, University of Chinese Academy of Sciences, Beijing, China. His research interests include robotics, SLAM, sensors fusion and deep learning.
\end{IEEEbiography}
\begin{IEEEbiography}
	[{\includegraphics[width=1in,height=1.25in,clip,keepaspectratio]{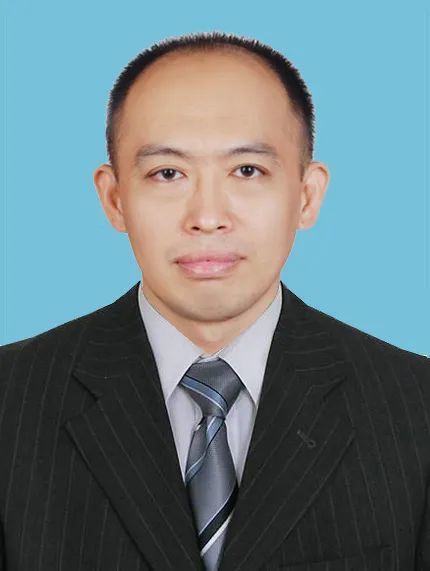}}] {Rui Guo}
	received his Ph.D. degree from the Harbin Institute of Technology, Harbin, China, in 2007.  
	He is currently the chief expert of the Electric Power Research Institute, State Grid Shandong Electric Power Company, Jinan, China. His current research interests include electric robots and power electronics technology.
\end{IEEEbiography}
\end{document}